%% file: main_draft_v2.tex
\pdfoutput=1

\documentclass[11pt,dvipsnames]{article}

\usepackage[]{acl}

\usepackage{times}
\usepackage{latexsym}

\usepackage[T1]{fontenc}

\usepackage[utf8]{inputenc}

\usepackage{microtype}

\renewcommand{\thefootnote}{\fnsymbol{footnote}}
\usepackage[symbol]{footmisc} 
\usepackage{lipsum}
\usepackage{hyperref} 
\usepackage{booktabs} 
\usepackage[normalem]{ulem} 

\SetLipsumParListSurrounders{\colorlet{oldcolor}{.}\color{gray}}{\color{oldcolor}}

\usepackage{todonotes}
\usepackage{multirow}
\usepackage{tikz}
\usepackage{pgf}
\usepackage{pgfplots}
\usepackage{xcolor}
\usepackage{lmodern}

\newcommand{\specialcell}[2][c]{%
  \begin{tabular}[#1]{@{}c@{}}#2\end{tabular}}

\newif\ifcomments
\commentstrue
\ifcomments
\usepackage[normalem]{ulem}
\definecolor{CMpurple}{rgb}{0.6,0.18,0.64}
\newcommand\cm[1]{\textcolor{CMpurple}{\textsf{\scriptsize[\textbf{CM\@:} #1]}}}
\newcommand\cmi[1]{\textcolor{CMpurple}{#1}}
\newcommand\cmm[1]{\marginpar{\raggedright\tiny\textcolor{CMpurple}{\textsf{{\bfseries CM\@:} #1}}}}
\newcommand\cms{\bgroup\markoverwith{\textcolor{CMpurple}{\rule[.4ex]{2pt}{0.8pt}}}\ULon}

\newcommand\sidenote[1]{\marginpar{\raggedright\tiny\textcolor{black}{\textsf{#1}}}}
\newcommand\pagemessage[1]{\marginpar{\raggedright\tiny\textcolor{blue}{\textbf{Key page message: }\textsf{#1}}}}

\else
\newcommand\cm[1]{}
\newcommand\cmi[1]{\ignorespaces}
\newcommand\cmm[1]{}
\newcommand\cms[1]{#1}

\newcommand\sidenote[1]{}
\newcommand\pagemessage[1]{}
\fi

\title{Detecting Label Errors by using Pre-Trained Language Models}

\author{Derek Chong\footnotemark[1] \\
  Stanford University \\
  \texttt{\small derekch@stanford.edu} \\\And
  Jenny Hong\footnotemark[1] \\
  Stanford University \\
  \texttt{\small jennyhong@cs.stanford.edu}\\ \\\And
  Christopher D. Manning \\
  Stanford University \\
  \texttt{\small manning@cs.stanford.edu} \\
 }

\begin{document}
\maketitle

\begin{abstract} 
We show that large pre-trained language models are inherently highly capable of identifying label errors in natural language datasets: simply examining out-of-sample data points in descending order of fine-tuned task loss significantly outperforms more complex error-detection mechanisms proposed in previous work. 
To this end, we contribute a novel method for introducing realistic, human-originated label noise into existing crowdsourced datasets such as SNLI and TweetNLP. 
We show that this noise has similar properties to real, hand-verified label errors, and is harder to detect than existing synthetic noise, creating challenges for model robustness.
We argue that human-originated noise is a better standard for evaluation than synthetic noise.
Finally, we use crowdsourced verification to evaluate the detection of real errors on IMDB, Amazon Reviews, and Recon, and confirm that pre-trained models perform at a 9--36\% higher absolute Area Under the Precision-Recall Curve than existing models.
\end{abstract}

\footnotetext[1]{Equal contribution.}
\renewcommand{\thefootnote}{\arabic{footnote}}

\section{Introduction}


Improving model performance in the presence of label errors comprises an area of active research \cite{song2022survey}. However, existing methods focus on label errors in training data. Although seldom acknowledged, evaluation label errors are at least as pernicious as training label errors: pervasive errors in commonly used NLP benchmarks have been found to destabilize model performance \cite{malik2011automatic, northcutt2021pervasive}. Such findings suggest that improving training methods does not preclude the need for improving the underlying data. We propose a simple method for using large, pre-trained language models (LLMs) to directly identify label errors for the purposes of correcting or removing them. 

\begin{figure}
    \centering
    \input{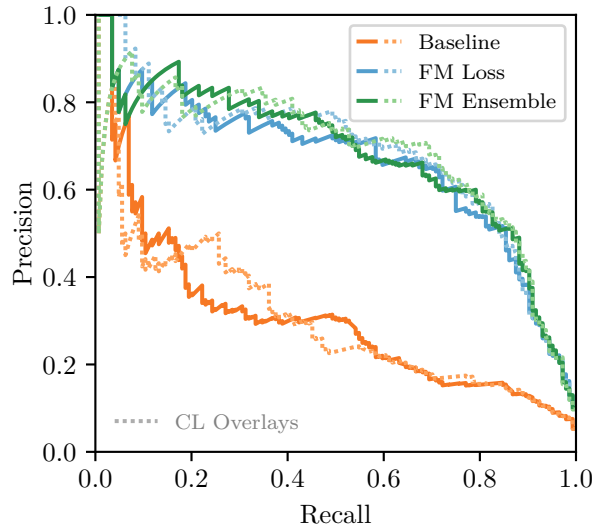}
    \caption{Precision-recall curves for label error detection: Large language models detect label errors with high precision, and far more effectively than a baseline word vector-based neural classifier. Overlaying a state-of-the-art model-agnostic error detection method, Confident Learning, results in little to no improvement (TweetNLP-5; \S\ref{sec:results}). }
    \label{fig:aupr_tweetnlp5_topleft}
\end{figure}

\input tables/error_examples_short

The majority of work in identifying label errors, and in data-centric artificial intelligence (DCAI) more broadly, focuses on image and healthcare data \cite{neurisdcai}.
However, the success of the foundation model (FM) paradigm in applying pre-trained language models to a variety of NLP tasks \citep{bommasani2021opportunities,reiss2020identifying} suggests that FMs may be a powerful tool for detecting and correcting label errors in language datasets.
Pre-training has been shown to imbue models with properties such as resistance to label errors, class imbalance \cite{karthik2021learning}, out-of-distribution detection \cite{hendrycks2018using}, and confidence calibration \cite{desai-durrett-2020-calibration}, while conferring robustness, generalization, and natural language understanding capabilities \cite{wang-etal-2018-glue, petroni-etal-2019-language}. Our primary contribution is to show that simply verifying items in order of their out-of-sample loss on a foundation model improves precision by an absolute 15--28\% and Area Under the Precision-Recall Curve (AUPR) by an absolute 9--36\%.

Many methods for label error detection rely on artificially introduced label errors as ground truth for evaluating their methods. \citet{northcutt2021confident} develop a state-of-the-art model for identifying label errors, Confident Learning (CL), and use the better approach of crowdsourced human evaluations to determine the ground truth of label errors. We model our experiments on real data after their verification protocol, replicating this on real errors in IMDB \cite{maas-etal-2011-learning}, Amazon Reviews \cite{amazonreviews}, and Recon \cite{hong2021nllp}, with  adaptations to mitigate annotator fraud \cite{kennedy2020shape}.

In the process of assessing our results, we contribute a novel technique and protocol for introducing realistic, human-originated label noise into existing crowdsourced datasets, and apply it to two such datasets, TweetNLP \cite{gimpel2010} and SNLI \cite{snli}.
We demonstrate that our technique better approximates \textit{organic} (real, naturally occurring) label errors than existing methods. We provide evidence that this realism is essential to properly assessing model performance: even models that are robust to standard synthetic noising approaches show limited robustness to human-originated noise.\footnote{Data noising library and evaluation data available at \href{https://github.com/dcx/lnlfm}{https://github.com/dcx/lnlfm}.}

\section{Related Work}

\label{sec:related_work}
Learning with Noisy Labels (LNL) focuses on the model-training stage.
Noise-robust approaches examine model enhancements such as the design of loss functions \cite{joulin2016learning, amid2019robust, liu2020peer, ma2020normalized}, regularization \cite{azadi2015auxiliary, zhou-chen-2021-learning}, reweighting \cite{bar2021multiplicative, kumar2021constrained}, hard negative mining and contrastive learning \cite{zhang-stratos-2021-understanding}. Noise-cleansing approaches aim to segregate clean data from noisy data in training, e.g. bagging and boosting \cite{wheway2000using, sluban2014ensemble},
$k$-nearest neighbors \cite{delany2012profiling},
outlier detection \cite{gamberger2000noise, thongkam2008support}, bootstrapping \cite{reed2014training},
and neural networks supervised directly on
detecting an error, when such data exist \cite{jiang2018mentornet}.

LNL methods have in most cases been evaluated using artificially-generated label noise.
A typical evaluation of an LNL method uses a standard benchmark dataset, and programmatically corrupts training labels via one of three main noising schemes \cite{frenay2014survey, algan2020label}.
\textit{Uniform noise} is most commonly used but unrealistic;
deep neural networks have been found to
perform well even when noised labels outnumber original labels at a ratio of 100 to 1 \cite{rolnick2017deep}.
\textit{Class-dependent noise} randomly permutes labels based on a confusion matrix.
However, research on annotator disagreement suggests that label errors tend to result from feature-based, not class-based ambiguity \cite{hendrycks2018using}. Training models to generate realistic \textit{feature-based} or \textit{instance-dependent noise} has recently emerged as an area of active research \cite{chen2021beyond, xu2021dataclue, dawson2021rethinking}. However, \citet{algan2020label} report that feature-dependent noise may bias benchmark performance toward similar models to the ones used to generate this noise. 

The noising schemes above each fail in some way to simulate organic, naturally occurring label errors, which are estimated to occur in common benchmarks at 1--5\% of labels  \cite{redman1998impact, muller2019identifying, northcutt2021pervasive, kreutzer2022quality} or even as much as 20\% \cite{hovy2014gimpel, abedjan2016detecting}.
For organic errors, CL \cite{northcutt2021confident} predicts errors in IMDB, Amazon Reviews, and other datasets by estimating a joint distribution between noisy and uncorrupted labels;
\citet{reiss2020identifying} pioneers using BERT for error detection on ConLL-2003 via a classifier trained over a frozen BERT embeddings layer. 

\section{Methods}

\paragraph{Motivation.}
Empirical evidence on image data suggests that models exhibit high loss on label errors in training data relative to the underlying features \cite{huang2019o2u, kim2021fine, hong2021nllp, chen2021detecting}. 
\citet{hendrycks17baseline} show that predicted probabilities of (non pre-trained) neural networks can identify out-of-distribution examples.
We consider the framing that label errors are one type of out-of-distribution data. 
Indeed, CL \cite{northcutt2021confident} uses normalized predicted probabilities, also from non pre-trained models, to directly identify label errors. Foundation models are highly performant; we hypothesize that a low likelihood label is likely to be an error.

\begin{figure}
    \centering
    \input{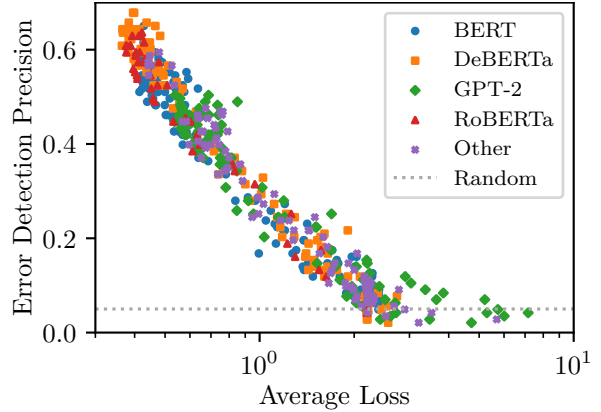}
    \caption{Loss exhibits a strong log-linear relationship with error detection precision at a fixed threshold, across a broad range of models and hyperparameters ($r^2$: 0.94; TweetNLP-5, \S\ref{sec:results-overall}).}
    \label{fig:loss_f1}
\end{figure}

\paragraph{Foundation models.}
\begin{figure*}
    \centering
    \input{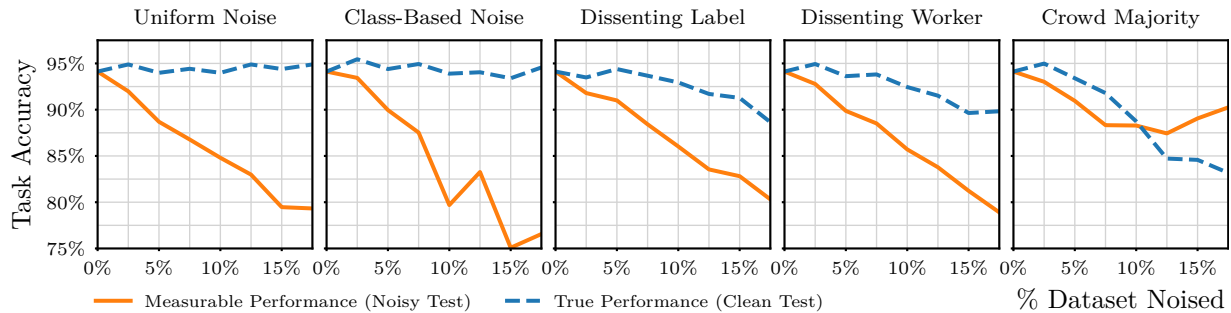}
    \caption{Assessing model robustness against a range of noising methods on TweetNLP, with methods ordered by hypothesized realism. Solid orange lines report task performance on noisy test data, reflecting observations in practice; dashed blue lines report task performance on underlying clean test data, reflecting models' actual performance. Models may be robust to uniform and class-dependent noise, where the true performance remains high even with increasing levels of noise. However, they are not necessarily robust to human-originated noise, where the true test performance decreases with increasing noise.}
    \label{fig:gimpel_testerr_multiplot}
\end{figure*}

The success of \citet{reiss2020identifying}'s approach in using frozen BERT embeddings motivates
directly applying the foundation model paradigm: we use a large language model that was first pre-trained on a task-agnostic dataset, then fine-tune the model for a given task. 

We address classification tasks: given a model's score $f_{i, c}$ for each item $i$ and class $c$, its predicted probability is the softmax-normalized score $p(c \mid x_i)$.
Because each item belongs to exactly one class, the contribution of item $i$ to the loss is the negative log probability of the score for the assigned class $y_i$:
$$
L_i = \sum_i - \textrm{log } p(y_i \mid x_i).
$$
We fine-tune such a model for the training split of each data set. To identify label errors on a validation or test set, we hypothesize items from the dataset as a label error in order of the item's loss on that out-of-distribution set.

We propose two main methods. Foundation Model Loss (FML) uses a single foundation model, fine-tuned on the corresponding task (e.g., sentiment classification, POS tagging), to hypothesize items in order of the model-predicted loss. We augment FML using task-adaptive pre-training (TAPT; \citealp{gururangan2020tapt}), which is further pre-training on in-domain data, using only text on the pre-training objective without using any labels for fine-tuning on the cross-entropy objective.

Foundation Model Ensembling (FME) combines multiple foundation models on the same task. 
We hypothesize that ensembling may be disproportionately effective at detecting label errors, as training noise induces models to learn random spurious correlations \cite{watson2022agree}.
Rather than using a validation set to choose the single model with the lowest loss on the task, FME uses the top three models trained in a hyperparameter sweep, and differing in both hyperparameters and random initialization, as fully described in Appendix~\ref{sec:appendix-modelselect}.
FME creates a synthetic probability distribution over the task outputs by averaging the probabilities predicted using each individual model. FME then hypothesizes items in order of loss over the synthetic distribution. 

\section{Generating Realistic Label Noise}
\label{sec:human-originated-noise}

To better evaluate label noise detection performance, we prepare a set of benchmark datasets populated with controllable, highly realistic, human-originated label noise. 

\paragraph{Sources of human error.}
We observe that datasets often undergo multiple annotation passes: crowdsourced labels typically aggregate several annotators' inputs \cite{hovy2014gimpel,wei2022learning}, and subsets of data may receive more extensive validation \cite{snli}, gold labels by trained experts \cite{plank2014linguistically}, or correction passes \cite{reiss2020identifying}.
We hypothesize that differences between such annotations may be usefully repurposed as a source of realistic, \textit{human-originated} label noise, as disagreements between annotators is known to reflect systematic ambiguity and human error \cite{plank2014linguistically,zhang2017consensus}, and differs from the type of noise studied using existing synthetic methods. 

We construct three noising methods which may be applied in many of the above scenarios. For any dataset which includes two levels of label quality, the \textit{dissenting label} method replaces final labels with disagreeing labels at random, simulating imperfect quality control. Datasets which provide individual annotator identifiers may apply the \textit{dissenting worker} approach: select one annotator at random, apply all of their labels which disagree with final labels, and repeat until reaching the target noise rate. This simulates gaps in annotator training, which introduce systematic idiosyncrasies. Finally the \textit{crowd majority} method applies to any dataset in which individual annotations can be aggregated to produce a label other than the final label: the former label simulates challenging, systematic errors in the latter.

\paragraph{Noising and robustness.} 
We assess the effect of these noising methods using TweetNLP \cite{gimpel2010}, a corpus of 26,435 tokens from 1,827 American English tweets collected from Twitter used to train part-of-speech (POS) tagging. TweetNLP includes gold labels annotated by 17 experts, but later received a separate crowdsourced assessment, aggregated by majority vote \cite{hovy2014gimpel}. We noise TweetNLP to eight levels from 0-20\% separately for each method, fine-tune \verb|DeBERTA-v3-base| \cite{he2021debertav3} on each noising, and evaluate models on both noisy and clean test sets. Results from noisy test sets represent model performance as \textit{measurable} in practice; real datasets contend with noise in evaluation data. Clean test set results represent \textit{true} model performance. Fig.~\ref{fig:gimpel_testerr_multiplot} reports the results of this evaluation.

\begin{figure}
    \centering
    \input{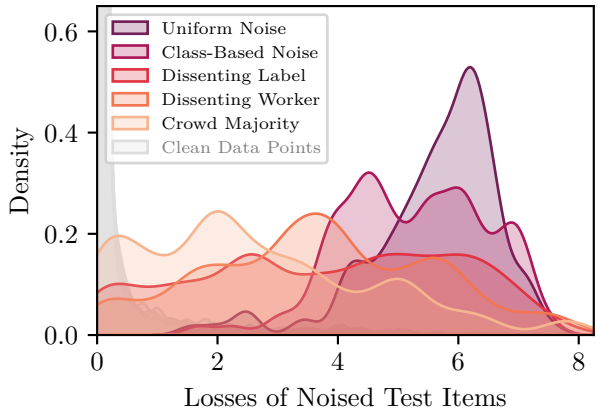}
    \caption{Distributions of losses of label errors on TweetNLP at 5\% noising. Uniform and class-based noise produce high and distinctive losses; human-originated noise is widely distributed, and has greater overlap with the distribution of clean data points; \S{6}.}
    \label{fig:loss_hist_noisyonly}
\end{figure}

For uniform and class-dependent noise, true performance remains high even for high noise levels (per \citealp{rolnick2017deep}).
But crucially, this robustness does not extend to human-originated noise: human label errors are correlated to input text, and so contain systematic erroneous features, which models may learn in training.
On more challenging noising methods, although measured performance appears to increase, true performance actually \textit{linearly decreases} with noise. 
Fig.~\ref{fig:loss_hist_noisyonly} explores this further via the distributions of model losses for each noising method:
loss induced by human-originated noise overlaps significantly with clean items, whereas loss from uniform and class-based noising is distinctively higher.

\paragraph{Noise detection benchmarks.}
We standardize a set of benchmarks from existing datasets for use in our main experiments. TweetNLP-5 and SNLI-5 aim to simulate typical data noise conditions: we apply dissenting worker and dissenting label noising to a 5\% level (see Appendix \ref{sec:appendix-noising} for details). SNLI is a corpus of 570,152 sentence pairs, in which the task is to label each pair with entailment, contradiction, or semantic independence; we use the 10\% subset which includes five crowdsourced annotations per item, as collected by \citet{snli} during data validation.

We construct TweetNLP-M to investigate robustness to systematic error introduced by the crowdsourcing process. 
We apply crowd majority noising, comparing noisy majority-vote aggregated labels by \citet{hovy2014gimpel}
to clean expert labels, which serve as a measure of true performance.
Accordingly, we retain all disagreements, or 20.46\% of the dataset. We also report results on Recon, a legal classification dataset of 1,279 documents in which \citet{hong2021nllp} found label errors to destabilize model evaluation; as above, we compare non-expert and expert annotator labels.

\section{Validation on Real Label Errors}
\label{sec:mturk-protocol}
In addition to human-originated noise datasets, we evaluate error detection performance on organic errors in two benchmark datasets, following \citet{northcutt2021confident}'s protocol.

\paragraph{Datasets.} 
The IMDB Large Movie Review Dataset is a collection of movie reviews for binary sentiment classification \citep{maas-etal-2011-learning}, and is split into train and test sets of 25,000 items each.
Amazon Reviews is a collection of reviews and 5-point star ratings from Amazon customers \citep{amazonreviews}.
We used the version released by \citet{northcutt2021confident}, which includes the following modifications:
It uses 1-star, 3-star, 5-star reviews with net positive helpful upvotes as a ternary sentiment task,
resulting in a dataset of 9,996,437 reviews.
For tractability we use a train split of a random sample of 2.5 million items, and a test split of 25,000 items.

\begin{table}[t]
\centering
\begin{tabular}{lcccc}
\specialrule{.1em}{0em}{0em}
\textbf{IMDB} & \multicolumn{4}{c}{New Protocol} \\\cline{2-5}
Old Protocol & C & NA & NE & Total \\\hline
Correctable & 105 & 44 & 24 & 173\\ 
Non-Agreement & 75 & 252 & 225 & \textcolor{red}{\textbf{552}} \\ 
Non-Error & 3 & 62 & 520 & 585 \\ \hline
Total & 183 & \textcolor{PineGreen}{\textbf{358}} & 769 & 1310 \\
\specialrule{.1em}{0em}{0em}
\\
\specialrule{.1em}{0em}{0em}
\textbf{Amazon} & \multicolumn{4}{c}{New Protocol} \\\cline{2-5}
Old Protocol & C & NA & NE & Total \\\hline
Correctable  & 142 & 43 & 117 & 302 \\ 
Non-Agreement & 140 & 79 & 211 & \textcolor{red}{\textbf{430}} \\ 
Non-Error & 75 & 31 & 162 & 268 \\ \hline
Total & 357 & \textcolor{PineGreen}{\textbf{153}} & 490 & 1000 \\
\specialrule{.1em}{0em}{0em}
\end{tabular}
\caption{Re-evaluation of baselines: The number of \textbf{C}orrectable, \textbf{N}on-\textbf{A}greement, and \textbf{N}on-\textbf{E}rror assessments produced by the CL Mechanical Turk evaluation protocol and the new protocol, on the same set of items. The new protocol substantially reduces annotator non-agreement; \S\ref{sec:mturk-protocol}.}
\label{tab:turk-eval-long}
\end{table}

\paragraph{Baseline protocol.}
Workers are presented with review text and asked to determine whether overall sentiment is positive, negative, neutral, or off-topic. Each review is independently presented to five workers. An example is considered a ``Non-Error'' if at least three workers agree the original label is correct. Otherwise, we consider the label to be correctly identified as an error. We further categorize label errors as ``Correctable'' if at least three workers agree on the same replacement label, or ``Non-Agreement'' if no majority exists. 

\paragraph{New adaptations.} While conducting initial experiments, we found that the \citet{northcutt2021confident} MTurk protocol resulted in
a significant amount of annotator fraud.
Some workers spent unreasonably short amounts of time on the text, and frequently disagreed with both expert and peer annotators, reflecting increasingly common issues in crowdsourced annotations \cite{kennedy2020shape}. 
Appendix~\ref{sec:appendix-mturk-protocol} describes four extra conditions we added to improve the \citet{northcutt2021confident} protocol.

In order to establish an accurate baseline, we re-evaluate the label errors hypothesized by CL \citep{northcutt2021confident}.
On the new protocol, Fleiss' $\kappa$ inter-annotator agreement increases from 0.131 to 0.464 for IMDB, and 0.014 to 0.556 for Amazon, and
Table~\ref{tab:turk-eval-long} shows that Non-Agreement decreases by 35\% in IMDB
and 65\% in Amazon.
This suggests a substantial decrease in low-quality annotations.

\section{Experiments} 
\label{sec:experiments}

\input{tables/aupr_full}

\begin{figure}
    \centering
    \input{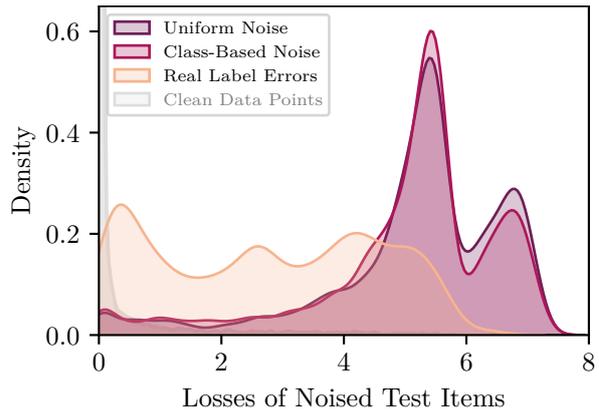}
    \caption{Distributions of losses of hypothesized label errors that MTurk workers verified for IMDB. As with Fig.~\ref{fig:loss_hist_noisyonly}, uniform and class-based methods do not approximate real, worker-identified errors, and losses of real label errors have greater overlap with the distribution of clean data; \S{6}.}
    \label{fig:loss_hist_realerrors}
\end{figure}

\paragraph{Label noise realism.}
Section~\ref{sec:human-originated-noise} defined the human-originated noising protocol used to generate TweetNLP-5, TweetNLP-M, and SNLI-5. Section~\ref{sec:mturk-protocol} specified a protocol for identifying organic label errors present in IMDB and Amazon. We assess the realism of synthetic noise methods by comparing loss distributions against models trained with organic noise (for real label errors, we refer to items verified as Correctable via MTurk). We quantify the degree to which noising induces erroneous learning by measuring the Wasserstein distances between noisy and clean loss distributions. 

\paragraph{Overall LLM performance.}
We assess broad error detection capabilities by evaluating 13 commonly-used LLMs on TweetNLP-5.
We measure performance against loss, model size, and GLUE score (a proxy for general model capability; \citealp{wang-etal-2018-glue}). Appendix~\ref{sec:appendix-modelselect} provides implementation details.
This experiment's results inform model selection: we use \verb|DeBERTA-v3-base| for all further experiments.\footnote{We also use RoBERTa-BigBird for Recon in order to handle its long input passages \cite{hong2021nllp}.}


\paragraph{Main experiment.}
Using our realistic noising benchmarks, and the MTurk baselines and verification protocol, we can now assess the performance of each label error detection method.
We evaluate Foundation Model Loss (\textbf{FML}) and Foundation Model Ensembling (\textbf{FME}).

As a baseline, we evaluate Confident Learning (\textbf{CL}; \citealp{northcutt2021confident}). CL is not a standalone method; it augments existing models. Given an underlying model's predicted scores for each class and the true proportion of each class, CL forms a reweighting matrix, called the confident joint. To form a label error prediction score, CL reweights the model's scores by the confident joint. CL hypothesizes items in order of this resulting score.

CL uses FastText \cite{joulin2017bag} for IMDB and Amazon, but includes no implementations for POS tagging or NLI. As a result, for TweetNLP and SNLI, we apply CL to the \textbf{H\&G} baseline \cite{hendrycks17baseline}, a two-layer neural classifier over
word vectors pre-trained on a corpus of 56 million tweets \cite{owoputi-etal-2013-improved}.
For all datasets, we also assess applying CL to foundation models (\textbf{FME+CL}).

For each dataset, we run 25 hyperparameter sweeps which each fine-tune a model for the given task (e.g., POS tagging) using noisy data, and select the model with the best validation set task performance.
We report label error detection performance (not task performance).
Area Under the Precsion-Recall Curve (AUPR) provides an overall performance score \cite{saito2015precision,hendrycks17baseline}. We also report metrics representing performance on competing data cleaning priorities: efficiency requires high precision on a small number of items, whereas coverage requires high recall on a larger number of items. Appendix~\ref{sec:appendix-metrics} describes the Truncated AUPR used for IMDB and Amazon, which are too costly to fully crowd verify. 

\stepcounter{footnote} \footnotetext{Precision and recall are equal when evaluating a number of items equal to the total error count.}

\paragraph{End-to-end noising.} 
We finally isolate the effects of noise and label error correction for validation and test splits.
For each dataset, we prepare three versions of the validation and test splits, respectively: a \textit{clean} version assumed to contain zero errors,\footnote{For TweetNLP, we justify our assumption in \S\ref{sec:human-originated-noise}: expert labels by \citet{hovy2014gimpel} are considered noise free compared to crowd labels. For IMDB and Amazon, we follow \citet{northcutt2021confident}, which adds several percentage points more noise than naturally occurs.}
a \textit{noisy} version, with label noise deliberately introduced,
and a \textit{corrected} version generated from noisy splits using our main error detection method (ranking errors with FME and correcting the top Err\% data points).
We train 40 hyperparameter sweeps, with performance cross-evaluated on all prepared data splits.

We report three different metrics.
We report each model's accuracy on the clean test split as the \textit{true} accuracy. Following the norms of Fig.~\ref{fig:gimpel_testerr_multiplot}, we report the \textit{measurable} accuracy as the accuracy of the model selected using performance on the noisy or corrected validation split on the corresponding test split.
Finally, we report the \textit{rank} of the model as the rank of the model's performance on clean test data. The best performing model among all sweeps has rank 1, and the worst has rank 40. This metric emphasizes that different validation sets select different models.

We perform this exercise using IMDB and Amazon noised to 5\% (I-5, A-5), and TweetNLP-5 and TweetNLP-M.

\section{Results}
\label{sec:results}

\begin{figure}
    \centering
    \input{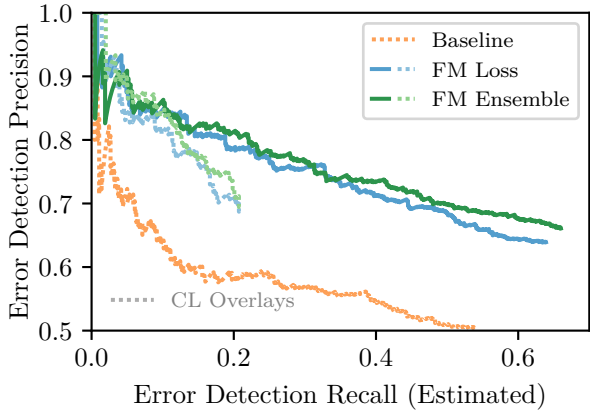}
    \caption{Precision-recall curves for label error detection on Amazon by method. FML+CL and FME+CL produce fewer items and do not extend to a recall past 0.21. Applying CL to FM changes little compared to using FM alone.}
    \label{fig:aupr_amazon}
\end{figure}

\paragraph{Label noise realism.}
Human-originated noise appears to closely approximate real label noise. Figs.~\ref{fig:loss_hist_noisyonly} and~\ref{fig:loss_hist_realerrors} show that the losses of both real and human-originated label errors are lower and more widely-distributed than existing noising methods. Their Wasserstein distances to the distribution of clean data are significantly lower than existing noising methods, suggesting comparable erroneous learning (Appendix~\ref{sec:appendix-lossdist}). 

\begin{figure}
    \centering
    \input{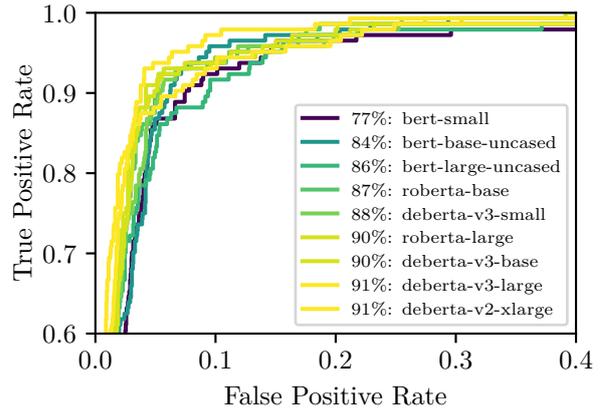}
    \caption{ROC curves for error detection performance on TweetNLP-5: LLM loss is highly effective for detecting label errors, and performance is highly correlated with general language understanding (GLUE, $r^2$: 0.79).
    }
    \label{fig:roc-v-nlu}
\end{figure}

\paragraph{Overall LLM performance.} \label{sec:results-overall}
We discover a strong log-linear relationship between error detection performance and loss, which holds across many model families and configurations ($r^2$: 0.94, Fig.~\ref{fig:loss_f1}).
We also find relationships between error detection performance and general model capability, in terms of GLUE score ($r^2$: 0.79) and model size (Fig.~\ref{fig:model_size_f1}). 
Fig.~\ref{fig:roc-v-nlu} illustrates key findings using models' receiver operating characteristic (ROC) curves.
Ensembling confers signficantly more gains in error detection performance higher than gains on underlying task performance, across a broad range of models and hyperparameters; Appendix~\ref{sec:appendix-ensembling} explores ensembling in greater detail.

\paragraph{Main experiment.} 
Table~\ref{tab:aupr_precision_recall_all} shows that Foundation Model Ensembling significantly improves AUPR from the CL and H\&G baselines on all datasets, with an absolute difference of 0.36 on IMDB, 0.09 on Amazon, and a difference of 0.07--0.44 on synthetic data. 

Fig.~\ref{fig:aupr_tweetnlp5_topleft} shows that applying CL to FME has minimal effect on performance at every level of recall; most numbers are identical across the FME and FME+CL rows of Table~\ref{tab:aupr_precision_recall_all}. In fact, CL does not necessarily improve upon the H\&G baseline across datasets, with CL performance sometimes dipping below H\&G by 0.01--0.03.

While loss naturally ranks all data points, CL only hypothesizes a fixed number of potential errors:
Appendix~\ref{sec:appendix-rawcounts} shows the raw counts of items at fixed thresholds, per the original CL study. At the CL threshold, we outperform CL by an absolute 15--28\%. At the CL+FME threshold, predicted items are almost exactly the same, with Jaccard similarities of 0.59--0.99. By contrast, ensembling improves performance over FML by a greater amount on almost every measure, and introduces no such constraint.

\paragraph{End-to-end noising.} 
Cleaning validation data selects better models.
Noise in validation splits reduces performance by encouraging the selection of models with lower true performance. Noise in test splits significantly reduces measureable (noisy test) performance, as expressed by the difference between measureable and true performance.
In general, correcting label errors improves task performance: even when the reported task performance worsens, the reported performance is closer to the true performance of the model, measured using clean training and validation data.

\begin{table}[]
\centering
\small
\begin{tabular}{llcccc}
\toprule
Eval. & Test Perf. & I-5 & A-5 & T-5  & T-M  \\ \midrule
\multirow{3}{*}{Noisy}   & Measurable & 90.1 & 88.3  & 89.3 & 89.3 \\
                         & True     & 94.2 & 91.0  & 92.8 & 82.0 \\
                         & Rank  & 10   & 1     & 3    & 10   \\ \midrule
\multirow{3}{*}{Corr.} & Measurable & \textcolor{PineGreen}{95.1} & \textcolor{PineGreen}{90.7}  & \textcolor{PineGreen}{92.9} & \textcolor{PineGreen}{88.5} \\
                         & True     & \textcolor{PineGreen}{95.1} & \textcolor{Red}{90.8}  & \textcolor{PineGreen}{93.0} & 82.0 \\
                         & Rank & \textcolor{PineGreen}{4}    & \textcolor{Red}{5}     & \textcolor{PineGreen}{2}    & \textcolor{PineGreen}{8}    \\ \midrule
Clean & True & 95.8 & 91.0  & 93.8 & 82.1 \\ 
\bottomrule
\end{tabular}
    \caption{End-to-end effects of label noise on task performance, as evaluated on noisy, corrected, and clean validation and test data splits. 
    True accuracy is measured on clean test sets, and measurable accuracy on noisy or corrected test sets. 
    Rank is a relative measure of true accuracy; lower numerical ranks have higher accuracy.
    Corrections which improve or reduce performance metrics are highlighted in green or red, respectively.
    Metrics are evaluated on models trained on noisy data. 
    }
    \label{tab:t5-cleanval-gain}
\end{table}

\section{Discussion}

\paragraph{Rapid data ``health check''.}
Sorting evaluation data by each item's loss is an easy way to quickly highlight label errors. 
Using this simple technique with a foundation model appears to generally identify over half of all label errors through human re-evaluation of a single-digit percentage of all data (Table~\ref{tab:aupr_precision_recall_all}).
We expect this technique to work across deep learning domains, due to its simplicity and the extensive use of training loss in LNL research \cite{song2022survey}. 
Given estimates for typical rates of label errors and the gain observed in the end-to-end experiment, our technique may enable a 1--2\% increase in reportable test accuracy across many datasets, in addition to the gains from improving model selection.

\paragraph{Pre-training and robustness.}
We demonstrate that despite established findings on artificial noising \cite{hendrycks2018using}, pre-training confers limited robustness to realistic human noise.
The majority of label errors are systematic in nature \cite{snow-etal-2008-cheap, plank2014linguistically, samuel2022dark}, and crowdsourced labels form, to an extent, a different distribution from reality, as approximated by expert labels \cite{hendrycks-etal-2020-pretrained}.
When trained on crowdsourced or other data containing systematic errors, FMs quickly drift towards this incorrect distribution.

\paragraph{Applying AI to data-centric AI. } 
Data-centric AI aims to improve AI through labeling, curating, and augmenting the underlying data. 
We find that AI itself can be applied towards improving data quality, as part of a human-in-the-loop (HITL) iteration, which contributes an additional positive feedback loop between data quality and AI performance.

\paragraph{New challenges in LNL.}
Standard noising methods are unrealistic and no longer challenging for state-of-the-art language models \cite{algan2020label}; recent LNL analyses study conditions where up to 80\% of labels are noised \cite{song2022survey}. Our findings reinforce the need to reassess LNL methods in the context of more realistic noise \cite{zhu2022bert}.

Our human-originated noising method produces realistic label errors, and can be applied to any crowdsourced dataset which includes raw annotation data. As such datasets emerge across deep learning domains \cite{wei2022learning}, we hope this method may inspire challenging and realistic new LNL performance benchmarks. Our method also enables detailed exploration of the properties of human noise, 
which may support work on open LNL problems such as improving feature-based noising techniques, and estimating dataset noise \cite{bauerle2022symphony, northcutt2021pervasive}.

\paragraph{End-to-end noising.} 
The study of model performance on noise in validation and test data is essential: noise in other splits can affect reported model performance as much as noise in training data.
Clean and noisy performance on evaluation data provide useful insight into models' overall performance.

\section{Conclusions and Future Work}

Pre-trained models effectively identify label errors on real NLP datasets, definitively outperforming existing methods on the same benchmarks by an absolute 9--36\% in AUPR. 

Human-originated noising techniques may present a solution to the clear limitations of current LNL noising schemes: they are highly realistic and yet controllable for experimental purposes. We invite further exploration of this family of label noising techniques. 
We believe human-originated noising enables future advancements across multiple areas of LNL, supporting new tasks and metrics in areas such as the cost of human reannotation, estimation of dataset error, and mitigation of bias. 

Finally, we advocate for LNL to move towards an end-to-end approach of \textit{evaluating with label noise}, which takes into account noise within validation and test splits, and more accurately models the conditions of data in practice. 

\section*{Limitations}

\paragraph{Partial metrics.}
Determining the true recall of a label error detection method on a real datasets is generally infeasible due to its high cost; this requires a complete re-evaluation so as to identify every label error within the dataset. While some datasets exist in which this has been undertaken, such as \citet{hovy2014gimpel} for TweetNLP, for most datasets containing organic label errors, we can only assess precision directly.

To mitigate this, we can estimate recall by estimating total dataset error counts using sampling techniques. As a result of this limitation, we prefer AUPR over AUROC (Area Under the Receiving Operating Curve) as our overall assessment metric: estimates of AUPR are scaled by a fixed ratio, and therefore comparable between models on the same dataset, whereas AUROC is nonlinear with respect to the estimate.

\paragraph{Requires multiple annotations per label.}
Human-originated noising methods are only applicable to datasets which include at least two human annotations per label. While it is becoming increasingly common to release individual-level annotator data, this is not an ubiquitous practice.

\paragraph{Cleaning benchmark data.} 
In our analysis of model performance gains derived from applying our methods to cleaning evaluation data, we find that cleaning validation splits enables the selection of models with better test performance. Such a method may be useful in a large number of applications.

However, we caution against using this method to clean data intended for use in comparing performance across model families and variants: the cleaning process may bias any such benchmarks toward the models most similar to the model used to clean the data. While our method improves the performance of a given model on a task, and correcting label errors always improves the validity of test data, these improvements is unlikely to improve the performance of all models by the same amount. 

This limitation is shared with other existing model-based scoring methods such as BERTScore \cite{bertscore}.

\section*{Acknowledgements}
We would like to thank Google.org for credits for use of the Google Cloud Platform.
We are also grateful to our anonymous reviewers, members of the Stanford NLP Group, and Bryan H. Chong for their constructive feedback, as well as the many researchers who made data publicly available to enable our present work.

\bibliography{anthology,custom}
\bibliographystyle{acl_natbib}

\appendix

\section{Noising Benchmarks}
\label{sec:appendix-noising}

This section specifies how noising protocols were applied to create each fixed crowdsourced dataset.
Crowd labels for each dataset are available to download from the respective GitHub projects.

\subsection{TweetNLP-5}
TweetNLP-5(\%) is a fixed noising of TweetNLP to a 5\% noise level in each split. Of the label errors, 80\% (i.e. 4\% of each split) are assigned using the \textit{dissenting worker} method. The remaining 20\% (i.e. 1\% of each split) are assigned using the \textit{dissenting label} method. Fig.~\ref{fig:loss_hist_noisyonly} shows that both methods provide similar distributions of label errors. Although the dissenting worker method more realistically captures individual worker idiosyncrasies, the dissenting label method is actually slightly lower loss during training (i.e. harder for a model to distinguish from correct labels).

\subsection{TweetNLP-M}
TweetNLP-M(ajority) directly uses the majority class labels collected by \citet{hovy2014gimpel} on the Crowdflower platform, which have a 79.54\% agreement with the high-quality expert gold labels collected by \citet{gimpel2010}. 
Per the \citet{hovy2014gimpel} protocol, in the rare case of ties, the tie is broken in favor of the label that matches the gold label, if applicable. Otherwise, a label is selected at random. The ``-M'' suffix distinguishes the \citet{hovy2014gimpel} labels from the gold labels. 
\subsection{SNLI-5}
The Stanford Natural Language Inference dataset (SNLI) annotations do not include a worker identifier, meaning each item is attached to five crowdsourced labels, but there is no indication of which labels came from the same annotator across the dataset. As a result, we cannot apply the dissenting worker noising method. 

SNLI-5 has exactly 5\% of its data noised in each split. Of the label errors, 80\% (i.e. 4\% of each split) are assigned using a method that represents systematic errors, to simulate of dissenting worker method: We use the minority label when there is a 3-2 split between the five labels. The remaining 20\% (i.e. 1\% of each split) are assigned using the dissenting label method, as in TweetNLP-5.

\begin{figure}[t]
    \centering
    \input{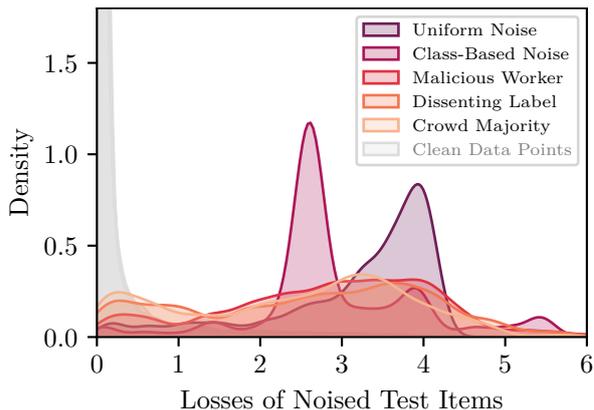}
    \caption{Distributions of losses of label errors on SNLI at 5\% noising, which demonstrates similar performance characteristics to TweetNLP, as shown in Fig.~\ref{fig:loss_hist_noisyonly}.}
    \label{fig:loss_hist_noisyonly_snli}
\end{figure}

\begin{figure}[t]
    \centering
    \input{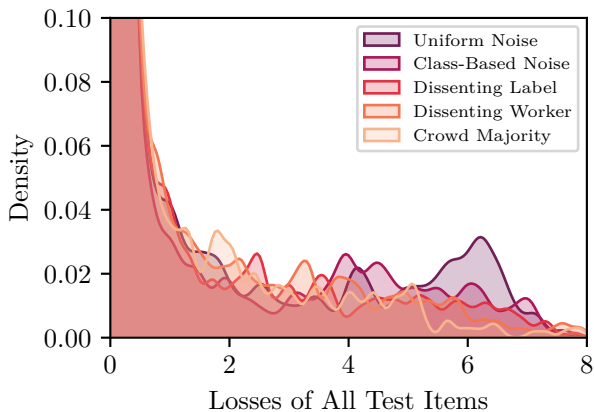}
    \caption{Combined distributions of losses of both noisy and clean data points, for TweetNLP with 5\% noising.}
    \label{fig:loss_hist_all}
\end{figure}

\section{Loss Distributions} 
\label{sec:appendix-lossdist}
Section~\ref{sec:human-originated-noise} examines dataset noisings primarily in terms of loss distributions on noised labels. To provide additional context, Fig.~\ref{fig:loss_hist_noisyonly_snli} provides an equivalent view for SNLI, and Fig.~\ref{fig:loss_hist_all} shows combined distributions of both clean and noisy data points on TweetNLP.

Table~\ref{tab:loss_wasserstein} reports the Wasserstein distances (or earth mover's distances) measured between the loss distributions of noisy and clean data points for models trained on TweetNLP and IMDB, as described in Section~\ref{sec:results}. Human-originated label noise more closely resembles both clean data points and real label noise as its hypothesized realism increases.

\begin{table}[]
\centering
\small
\begin{tabular}{lcc}
\toprule
Noising Method    & TweetNLP & IMDB \\ \midrule
Uniform Noise     & 5.62     & 5.04 \\ 
Class-Based Noise & 5.23     & 4.91 \\ \midrule
Dissenting Label  & 4.02     & - \\
Dissenting Worker & 3.44     & - \\
Crowd Majority    & \textbf{2.33}     & - \\ \midrule
Real Label Errors & -        & \textbf{2.67} \\
\bottomrule
\end{tabular}
\caption{Wasserstein distances between loss distributions of noisy and clean data points: Human-originated noising exhibits comparable levels of erroneous learning to organic label errors.}
\label{tab:loss_wasserstein}
\end{table}

\section{Mechanical Turk Protocol} 
\label{sec:appendix-mturk-protocol}

\begin{table}
\centering
\begin{tabular}{lcc} 
\toprule
 & IMDB & Amazon \\
\midrule
Original Protocol & 0.1314 & 0.0141 \\ 
New Protocol & \textbf{0.4643} & \textbf{0.5561}  \\ 
\bottomrule
\end{tabular}
\caption{A comparison of inter-annotator agreement between the original and new MTurk protocol results using Fleiss' $\kappa$. A score of 1.0 represents perfect agreement between workers, and 0.0 represents guessing at random. Annotations from the original protocol are substantially closer to random chance.}
\label{tab:turk-eval-agreement}
\end{table}

\subsection{Change Specifications}

We use Amazon Mechanical Turk to validate real label errors from IMDB \cite{maas-etal-2011-learning} and Amazon Reviews \cite{amazonreviews}.
We begin with the \citet{northcutt2021confident} protocol, and add four additional conditions, so as to mitigate annotator fraud.

First, we pre-qualify workers by requiring them to correctly answer a qualification test of four unambiguous questions \cite{hovy2014gimpel, agley2021quality}.

Second, after the initial qualification, we continue to monitor worker quality by introducing sentinel questions with known answers into the workers' regular tasks. We periodically remove workers who fail the tasks.

Third, we set filter criteria to limit workers to the following Anglosphere countries: United States, Canada, United Kingdom, Ireland, Australia, and New Zealand \cite{mturkbots}, to improve the chances of finding annotators with sufficient cultural context to correctly interpret review text.\footnote{Despite these precautions, we recognize that every precaution is subject to fraud, e.g., location is subject to VPN and bot attacks. \cite{dennis2020online, mellis2020mechanical, kennedy2020shape}} Our filter criteria include the standard recommendations of requiring a $\geq$ 99\% positive task approval rate with $\geq$ 500 tasks approved. 

Finally, we set a baseline target rate of US\$10 per hour, calculated using word counts and average reading speed (primarily for ethical reasons; the effect of compensation and annotation quality is an area of active research;  \citealp{saravanos2021hidden}).

The new protocol's labels are produced using a final set of approximately 70 workers. 
Workers averaged at least 12 seconds on each task; half the time needed to read prompts at an average reading speed.
The average time spent by a worker in the \citet{northcutt2021confident} protocol was 5 seconds.\footnote{The reported time is an \textit{upper} bound on the average time a worker spends on a task.}

\subsection{Protocol Validation}
We hypothesize that the Non-Agreements in the original protocol represent not only ambiguous data points, but also noise in the original protocol resulting from low quality work. Tables~\ref{tab:turk-eval-long} and~\ref{tab:turk-eval-agreement} show that the new protocol improves the level of agreement between workers. As such, we confirm that the increased agreement between workers in the new protocol results from higher quality labels.

Following the \citet{northcutt2021confident} protocol for expert review, we additionally select a total of 50 items from each of IMDB and Amazon for expert review. The experts are blinded to both the original labels and MTurk results and asked to label each item from scratch. They then reconciled results and came to a consensus for each item. The results are compared at the aggregate level of ``Correctable,'' ``Non-Agreement,'' and ``Non-Error,'' as opposed to the individual sentiment level (Positive, Negative, Neutral, or Off-Topic). The expert agreement with one another was 79\%, so in 21\% of the items, the expert label was considered to be Non-Agreement and matched the MTurk workers only if the workers also produced Non-Agreement. Table~\ref{tab:turk-eval-short} provides the result of this assessment.

For the original protocol, 52\% of the items agreed with expert annotators, 31\% of the items were incorrectly labeled as Non-Agreement, 12\% of the items were incorrectly labeled as Correctables, and 5\% of the items were incorrectly labeled as Non-Errors. 8\% of items were disagreements between experts and crowd workers where neither side had a Non-Agreement. In other words, 8\% of all items were disagreements between Correctable and Non-Error. 

For the new protocol, 72\% of the items agreed with expert annotators, 4\% of the items were incorrectly labeled as Non-Agreement, 7\% of the items were incorrectly labeled as Correctable, and 17\% were incorrectly labeled as Non-Errors. 5\% of items were disagreements between experts and crowd workers where neither side had a Non-Agreement.

\begin{table}
\centering
\small
\begin{tabular}{lccc}
\toprule
 & IMDB & Amazon & Total \\
\midrule
Original Correct & 33 & 19 & 52 \\ 
New Correct & \textbf{41} & \textbf{31} & 72 \\ 
Both Correct  & 28 & 14 & 42 \\ 
\bottomrule
\end{tabular}
\caption{A comparison of original and new MTurk protocol results against 100 expert-labeled data points.}
\label{tab:turk-eval-short}
\end{table}

\begin{table*}
\centering
\small
\begin{tabular}{lcccccccccc}
\toprule
\multirow{2}{*}[-2pt]{Dataset} & \multirow{2}{*}[-2pt]{\specialcell{Num. Errors\\Hypothesized}} & \multicolumn{3}{c}{Correctable} & \multicolumn{3}{c}{Non-Agreement} & \multicolumn{3}{c}{Non-Error} \\
\cmidrule(lr){3-5}\cmidrule(lr){6-8}\cmidrule(lr){9-11}
& & CL  & FML & FME & CL  & FML & FME & CL  & FML & FME \\ \midrule
IMDB & 1310 & 183 & 323 & \textbf{328} & 358 & 573 & 581 & 769 & 414 & 401 \\ 
Amazon & 1000 & 357 & 508 & \textbf{517} & 148 & 131 & 143 & 495 & 361 & 340 \\ 
TweetNLP-M & 250 & 121 & 158 & \textbf{165} & - & - & - & 129 & 92 & 85 \\ 
\bottomrule
\end{tabular}
\caption{The number of each type of error accurately identified for each dataset by each noise detection method, keeping the number of errors hypothesized fixed for ease of comparison. (TweetNLP is expert reviewed and by construction does not have any Non-Agreement types.)}
\label{tab:t3long-correctable}
\end{table*}

\begin{table*}
\centering
\small
\begin{tabular}{lcccccccc}
\toprule
\multirow{2}{*}[-2pt]{Dataset} & \multirow{2}{*}[-2pt]{\specialcell{Num. Errors\\Hypothesized}} & \multicolumn{2}{c}{Correctable} & \multicolumn{2}{c}{Non-Agreement} & \multicolumn{2}{c}{Non-Error} & \multirow{2}{*}[-2pt]{\specialcell{Jaccard\\Similarity}} \\
\cmidrule(lr){3-4}\cmidrule(lr){5-6}\cmidrule(lr){7-8}
& & FME & FME+CL & FME & FME+CL & FME & FME+CL \\ \midrule
IMDB & 316 & 168 & 168 & 108 & 108 & 40 & 40 & 0.99 \\ 
Amazon & 381 & \textbf{226} & 204 & 65 & 56 & 90 & 121 & 0.60 \\ 
TweetNLP-M & 129 & 93 & \textbf{98} & - & - & 36 & 31 & 0.59 \\ 
\bottomrule
\end{tabular}
\caption{Examining the performance of overlaying Confident Learning on FME, comparing the number of errors hypothesized by FME+CL. We also report the Jaccard similarity between the two models.}
\label{tab:t4long-correctable}
\end{table*}

\section{Overall LLM Performance Experiments}
\label{sec:appendix-modelselect}

Due to the high costs associated with expert and crowdsourced validation, we use TweetNLP-5 as a development dataset for model selection. 

We selected the following models for exploration: XLNet (\verb|base|, \verb|large|), RoBERTa (\verb|base|, \verb|large|), BERT (\verb|small|, \verb|base|, \verb|large|), DeBERTa (V3: \verb|xsmall|, \verb|small|, \verb|base|, \verb|large|, and V2: \verb|xlarge|, \verb|xxlarge|), GPT (assorted).
We performed 25 hyperparameter sweeps with each model, selecting the top three runs for further analysis. In order to avoid model family-level bias in the choice of hyperparameters, we set a broad shared range for three hyperparameters: learning rate varying from $10^{-6}$ to $10^{-3}$, the number of epochs from 2 to 8, and the batch size between 8, 16, 64, and 128. Training time and the final hyperparameters varied based on the model.

We ultimately selected \verb|DeBERTA-v3-base| as a compromise between performance and training speed. 
We used Google Cloud Platform for training infrastructure. Experiments were run using NVIDIA A100 GPUs, and runtime per training run was approximately 20 minutes for IMDB, Recon, and SNLI, 3 minutes for TweetNLP, and 4 hours for Amazon, when configured with a 2.5 million data point training split.

\begin{figure}
    \centering
    \input{fig/D04_model_size_vs_performance_f1_ensembled_probs_vs_cl.pgf}
    \caption{Label noise detection performance by model size and family, evaluated on TweetNLP-5. GPT-based models exhibit similar scaling trends, despite intrinsic disadvantages on classification tasks (due to pure autoregressive pre-training).}
    \label{fig:model_size_f1}
\end{figure}
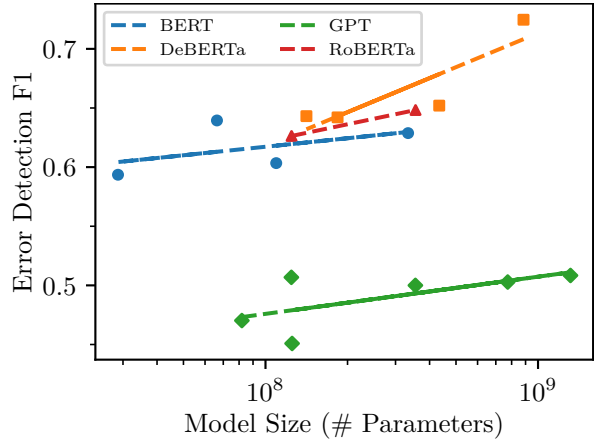

\section{Main Experiment}
\label{sec:appendix-mainexp}

\subsection{Metrics}
\label{sec:appendix-metrics}
We calculate the Area Under the Precsion-Recall Curve (AUPR) using the trapezoidal rule, given individual measurements of precision and recall at every possible threshold. 

We report the Truncated AUPR on IMDB and Amazon. Because IMDB and Amazon are too expensive to fully crowd verify, we cannot calculate precision and recall at the 25,000th item for each method, for each dataset, as it would require every data point to be relabeled on MTurk. Instead, we use the CL framework of predicting a fixed number of items. For example, for IMDB, CL hypothesizes 1,310 out of the 25,000 items to be label errors. We can calculate the precision and recall for every threshold, up to the number hypothesized by Confident Learning.  We can calculate the precision and recall of the 1st, 2nd, 3rd, \ldots, and 1,310th items.

We know the exact recall for all synthetic datasets. For IMDB and Amazon, we use the estimate that 5\% of the data is erroneous, which is consistent with common understanding of the prevalence of label errors \cite{redman1998impact, muller2019identifying, northcutt2021pervasive, kreutzer2022quality}.

All results reported on synthetic datasets reflect the average of individual scores from the three top-performing models from 25 hyperparameter sweeps. However, for cost-efficiency, results which require crowdsourced evaluation (such as IMDB and Amazon) are based on one run selected at random from a top three.

\subsection{Confident Learning}
\label{sec:appendix-rawcounts}
\citet{northcutt2021pervasive} reports results using raw counts, not the accuracy, precision, recall, or any other metric. For ease of comparability, Table~\ref{tab:t3long-correctable} reports the number of correctable, non-agreement, and non-error items identified by each method on each dataset. CL hypothesizes a fixed number of items, which is reported in the last column, and we assess a matching number of items from each method. 

When hypothesizing a fixed number of items, the foundation model approaches far outperform CL baselines. On IMDB, FME correctly identifies 909 label errors, a 28\% absolute improvement in accuracy. On Amazon, the FME approach correctly identifies 660 label errors, compared to the 505 identified by CL, a 15.5\% absolute improvement.

Applying CL to FME results in a different model that hypothesizes a different number of items (fewer, in all cases).
Table~\ref{tab:t4long-correctable} shows the raw counts of correctable, non-agreement, and non-error items when each of our models hypothesizes items at this reduced threshold.

\begin{table*}
\small
\centering
\setlength{\tabcolsep}{3pt}
\begin{tabular}{lcccccccc}
\toprule
\multirow{2}{*}[-2pt]{\specialcell{Method}} & \multicolumn{2}{c}{Task Accuracy} & \multicolumn{3}{c}{FM Error Detection Performance} & \multicolumn{3}{c}{Effects of CL Overlay} \\
\cmidrule(lr){2-3}\cmidrule(lr){4-6}\cmidrule(lr){7-9}
& Noisy & Clean & Precision & Recall & F1 & Precision & Recall & F1 \\
\midrule
Averaged & $0.88\pm0.03$ & $0.91\pm0.03$ & $0.50\pm0.11$ & $0.65\pm0.03$ & $0.56\pm0.08$ & $0.51\pm0.11$ & $0.67\pm0.03$ & $0.57\pm0.07$ \\ 
Ensembled & $0.89\pm0.02$ & $0.92\pm0.03$ & $0.56\pm0.12$ & $0.62\pm0.03$ & $0.58\pm0.08$ & $0.58\pm0.11$ & $0.65\pm0.03$ & $0.61\pm0.07$ \\ \midrule
Difference & $+1.14\%$ & $+1.24\%$ & $+12.52\%$ & $-4.31\%$ & $+4.66\%$ & $+13.89\%$ & $-2.98\%$ & $+6.03\%$ \\
\bottomrule
\end{tabular}
\caption{Ensembling confers gains in error detection performance disproportionate to gains in underlying task performance, across a broad range of models and hyperparameters (on TweetNLP-5, results from top three models per sweep, as measured at the fixed threshold set by CL).}
\label{tab:dev-ensembling-long}
\end{table*}
Overlaying CL on foundation model loss appears to have little marginal utility. Table~\ref{tab:t4long-correctable} also shows a high Jaccard similarity across all datasets, suggesting that applying CL on top of an FM changes little about the items hypothesized. On many datasets, FME and FME+CL perform almost identically in the number of items correctly hypothesized, slightly harming performance on Amazon Reviews, and slightly improving it on TweetNLP-5 (Table~\ref{tab:dev-ensembling-long}). FME+CL decreases the total number of hypothesized items compared to FME because of the threshold set by CL. We compare the FME and FME+CL approaches at the reduced number of hypothesized items in order to assess the impact of CL in the presence of pre-training.

Not only is aggregate performance nearly identical, we see in Figs.~\ref{fig:aupr_tweetnlp5_topleft} and~\ref{fig:aupr_amazon} that FME and FME+CL perform similarly for the \textit{entire range} of items hypothesized along the Precision-Recall curve. The primary difference is that FME can continue hypothesizing items even past FME+CL's threshold. 

\subsection{Ensembling}
\label{sec:appendix-ensembling}
Results from Tables~\ref{tab:aupr_precision_recall_all} and \ref{tab:dev-ensembling-long} show that ensembling (FME) improves error detection performance over using a single model (FML) in almost every scenario tested, at a rate several times higher than gains to underlying task performance. 

We also observe a phenomenon of disproportionately high variance in model error detection performance: Table~\ref{tab:dev-ensembling-long} quantifies the standard deviation of the former at three times the standard deviation of performance on the underlying task, and Fig.~\ref{fig:loss_f1} shows this to be the case even when comparing models with a fixed loss. This finding persisted even when holding all hyperparameters and data constant, with only the random seed being changed. 

We hypothesize that label noise in training data induces models to learn spurious correlations, which cause models to make errors in a structured manner \cite{watson2022agree,jiang2021assessing}; this results in greater levels of model disagreement, with minimal impact on top-line performance. Ensembling may be disproportionately effective because it serves an added function of reducing variance caused by these low-quality features.

\subsection{TAPT}
We perform Task-Assisted Pretraining (TAPT; \citealp{gururangan2020tapt}) using the original hyperparameters everywhere except for the optimizer, in which we use AdamW instead of Adam for DeBERTa. 
We run TAPT on the all splits of the corresponding data for all datasets except Amazon Reviews, where because of its size, we use TAPT on only 50,000 data points, or 0.5\% of the full dataset. 
After running TAPT, we then run 25 fine-tune sweeps.

\end{document}

%% file: tables/error_examples_short.tex
\begin{table*}[ht!]
\centering
\small
\renewcommand{\arraystretch}{1.2}
\begin{tabular}{lp{10cm}ll}
\toprule
Dataset & Text & Label & Sentiment \\
\midrule
IMDB & It is really unfortunate that a movie so well produced {\color{Maroon} turns out to be such a disappointment.} I thought this was full of (silly) cliches. It had all sorts of differences that it tried to tie together (not a bad thing in itself) but the result is at best awkward, but in fact ridiculous--too many clashes that wouldn't really happen. Then {\color{Maroon} the end of the movie--the last 10 minutes--ruined all the rest.} At first I thought Xavier was OK but with retrospect I think he was pretty bad. And that's all really too bad, because technically it was really good, and the soundtrack was great too. So the form was good, but {\color{Maroon} the content pretty horrible.} & \color{blue}\sout{Positive} & \color{Maroon}Negative \\
IMDB & The ending made my heart jump up into my throat. I proceeded to leave the movie theater a little jittery. After all, it was nearly midnight. {\color{blue}The movie was better than I expected.} I don't know why it didn't last very long in the theaters or make as much money as anticipated. {\color{blue} Definitely would recommend.} & \color{Maroon}\sout{Negative} & \color{blue}Positive \\ 
Amazon & The new design {\color{Maroon}only has a thin layer} of cellulose sponge material.  It will not last as long. Already {\color{Maroon}showing signs of wearing out}. The picture {\color{Maroon}does not represent the item received.} & \sout{Neutral} & \color{Maroon}Negative \\
\bottomrule
\end{tabular}
\caption{Organic label errors from sentiment datasets IMDB and Amazon, shown with the original dataset label. Each example was hypothesized by our model to be erroneous, and later verified by crowd workers.}
\label{detailed-ea-edu}
\end{table*}

%% file: tables/aupr_full.tex
\begin{table*}[ht!]
\small
\centering
\begingroup
\setlength{\tabcolsep}{3.3pt} 
\begin{tabular}{lcccccccccccccccc}
\toprule
& \multicolumn{6}{c}{Area Under Precision-Recall Curve} & \multicolumn{6}{c}{Precision, Recall @ Error\%\footnotemark[3]} & \multicolumn{4}{c}{Recall @ 2 $\cdot$ Error\%} 
\\\cmidrule(lr){2-7}\cmidrule(lr){8-13}\cmidrule(lr){14-17}
& I & Am. & R & T-5 & T-M & S-5 & I & Am. & R & T-5 & T-M & S-5 & R & T-5 & T-M & S-5  \\\midrule
H\&G & - & - & - & 0.30 & 0.41 & 0.20
  & - & - & - & 0.31 & 0.44 & 0.22
  & - & 0.54 & 0.63 & 0.34
\\ 
CL & 0.24 & 0.31 & 0.25 & 0.30 & 0.41 & 0.17
   & 0.41 & 0.51 & 0.31 & 0.36 & 0.44 & 0.18
   & 0.46 & 0.47 & 0.63 & 0.32
\\ 
FML & 0.58 & 0.39 & 0.37 & 0.66 & \textit{0.48} & 0.54
    & 0.68 & 0.64 & \textbf{0.46} & 0.65 & 0.47 & 0.45
    & 0.62 & 0.88 & 0.64 & 0.66
\\ 
FME & \textbf{0.60} & \textbf{0.40} & \textbf{0.38} & \textit{0.68} & \textit{0.48} & 0.61
    & \textbf{0.69} & \textbf{0.66} & 0.38 & 0.66 & \textit{0.48} & 0.46
    & \textit{0.69} & 0.88 & 0.65 & \textit{0.68}
\\ 
FME+CL & 0.20 & 0.17 & 0.37 & \textit{0.68} & \textit{0.48} & \textbf{0.62}
       & - & - & 0.38 & \textbf{0.69} & \textit{0.48} & \textbf{0.47}
       & \textit{0.69} & \textbf{0.89} & \textbf{0.66} & \textit{0.68}
\\\bottomrule
\end{tabular}
\endgroup

\caption{Main experiment: Evaluating label error detection methods using datasets containing highly-realistic label errors (\textbf{I}MDB, \textbf{Am}azon Reviews, \textbf{R}econ, \textbf{T}weetNLP-\textbf{5}, \textbf{T}weetNLP-\textbf{M}, \textbf{S}NLI-\textbf{5}). Foundation model-based methods significantly outperform baselines on every dataset, as shown by an overall performance metric (AUPR). In practice, estimating the number of dataset errors and checking this many items quickly catches up to 69\% of errors, at the same accuracy (P,R@Err\%).\footnotemark[3] For improved coverage, checking twice this number of items catches up to 89\% of errors (R@2$\cdot$Err\%).}

\label{tab:aupr_precision_recall_all}
\end{table*}

%% file: fig/D04_model_size_vs_performance_f1_ensembled_probs_vs_cl.pgf
\begingroup%
\makeatletter%
\begin{pgfpicture}%
\pgfpathrectangle{\pgfpointorigin}{\pgfqpoint{3.032090in}{2.274068in}}%
\pgfusepath{use as bounding box, clip}%
\begin{pgfscope}%
\pgfsetbuttcap%
\pgfsetmiterjoin%
\definecolor{currentfill}{rgb}{1.000000,1.000000,1.000000}%
\pgfsetfillcolor{currentfill}%
\pgfsetlinewidth{0.000000pt}%
\definecolor{currentstroke}{rgb}{1.000000,1.000000,1.000000}%
\pgfsetstrokecolor{currentstroke}%
\pgfsetdash{}{0pt}%
\pgfpathmoveto{\pgfqpoint{0.000000in}{0.000000in}}%
\pgfpathlineto{\pgfqpoint{3.032090in}{0.000000in}}%
\pgfpathlineto{\pgfqpoint{3.032090in}{2.274068in}}%
\pgfpathlineto{\pgfqpoint{0.000000in}{2.274068in}}%
\pgfpathlineto{\pgfqpoint{0.000000in}{0.000000in}}%
\pgfpathclose%
\pgfusepath{fill}%
\end{pgfscope}%
\begin{pgfscope}%
\pgfsetbuttcap%
\pgfsetmiterjoin%
\definecolor{currentfill}{rgb}{1.000000,1.000000,1.000000}%
\pgfsetfillcolor{currentfill}%
\pgfsetlinewidth{0.000000pt}%
\definecolor{currentstroke}{rgb}{0.000000,0.000000,0.000000}%
\pgfsetstrokecolor{currentstroke}%
\pgfsetstrokeopacity{0.000000}%
\pgfsetdash{}{0pt}%
\pgfpathmoveto{\pgfqpoint{0.453704in}{0.415123in}}%
\pgfpathlineto{\pgfqpoint{3.032090in}{0.415123in}}%
\pgfpathlineto{\pgfqpoint{3.032090in}{2.274068in}}%
\pgfpathlineto{\pgfqpoint{0.453704in}{2.274068in}}%
\pgfpathlineto{\pgfqpoint{0.453704in}{0.415123in}}%
\pgfpathclose%
\pgfusepath{fill}%
\end{pgfscope}%
\begin{pgfscope}%
\pgfpathrectangle{\pgfqpoint{0.453704in}{0.415123in}}{\pgfqpoint{2.578386in}{1.858944in}}%
\pgfusepath{clip}%
\pgfsetbuttcap%
\pgfsetroundjoin%
\definecolor{currentfill}{rgb}{0.121569,0.466667,0.705882}%
\pgfsetfillcolor{currentfill}%
\pgfsetlinewidth{1.003750pt}%
\definecolor{currentstroke}{rgb}{0.121569,0.466667,0.705882}%
\pgfsetstrokecolor{currentstroke}%
\pgfsetdash{}{0pt}%
\pgfsys@defobject{currentmarker}{\pgfqpoint{-0.024056in}{-0.024056in}}{\pgfqpoint{0.024056in}{0.024056in}}{%
\pgfpathmoveto{\pgfqpoint{0.000000in}{-0.024056in}}%
\pgfpathcurveto{\pgfqpoint{0.006380in}{-0.024056in}}{\pgfqpoint{0.012499in}{-0.021522in}}{\pgfqpoint{0.017010in}{-0.017010in}}%
\pgfpathcurveto{\pgfqpoint{0.021522in}{-0.012499in}}{\pgfqpoint{0.024056in}{-0.006380in}}{\pgfqpoint{0.024056in}{0.000000in}}%
\pgfpathcurveto{\pgfqpoint{0.024056in}{0.006380in}}{\pgfqpoint{0.021522in}{0.012499in}}{\pgfqpoint{0.017010in}{0.017010in}}%
\pgfpathcurveto{\pgfqpoint{0.012499in}{0.021522in}}{\pgfqpoint{0.006380in}{0.024056in}}{\pgfqpoint{0.000000in}{0.024056in}}%
\pgfpathcurveto{\pgfqpoint{-0.006380in}{0.024056in}}{\pgfqpoint{-0.012499in}{0.021522in}}{\pgfqpoint{-0.017010in}{0.017010in}}%
\pgfpathcurveto{\pgfqpoint{-0.021522in}{0.012499in}}{\pgfqpoint{-0.024056in}{0.006380in}}{\pgfqpoint{-0.024056in}{0.000000in}}%
\pgfpathcurveto{\pgfqpoint{-0.024056in}{-0.006380in}}{\pgfqpoint{-0.021522in}{-0.012499in}}{\pgfqpoint{-0.017010in}{-0.017010in}}%
\pgfpathcurveto{\pgfqpoint{-0.012499in}{-0.021522in}}{\pgfqpoint{-0.006380in}{-0.024056in}}{\pgfqpoint{0.000000in}{-0.024056in}}%
\pgfpathlineto{\pgfqpoint{0.000000in}{-0.024056in}}%
\pgfpathclose%
\pgfusepath{stroke,fill}%
}%
\begin{pgfscope}%
\pgfsys@transformshift{1.390458in}{1.441123in}%
\pgfsys@useobject{currentmarker}{}%
\end{pgfscope}%
\begin{pgfscope}%
\pgfsys@transformshift{2.073570in}{1.597746in}%
\pgfsys@useobject{currentmarker}{}%
\end{pgfscope}%
\begin{pgfscope}%
\pgfsys@transformshift{0.570904in}{1.380373in}%
\pgfsys@useobject{currentmarker}{}%
\end{pgfscope}%
\begin{pgfscope}%
\pgfsys@transformshift{1.083505in}{1.663754in}%
\pgfsys@useobject{currentmarker}{}%
\end{pgfscope}%
\end{pgfscope}%
\begin{pgfscope}%
\pgfpathrectangle{\pgfqpoint{0.453704in}{0.415123in}}{\pgfqpoint{2.578386in}{1.858944in}}%
\pgfusepath{clip}%
\pgfsetbuttcap%
\pgfsetroundjoin%
\definecolor{currentfill}{rgb}{1.000000,0.498039,0.054902}%
\pgfsetfillcolor{currentfill}%
\pgfsetlinewidth{1.003750pt}%
\definecolor{currentstroke}{rgb}{1.000000,0.498039,0.054902}%
\pgfsetstrokecolor{currentstroke}%
\pgfsetdash{}{0pt}%
\pgfsys@defobject{currentmarker}{\pgfqpoint{-0.024056in}{-0.024056in}}{\pgfqpoint{0.024056in}{0.024056in}}{%
\pgfpathmoveto{\pgfqpoint{-0.024056in}{-0.024056in}}%
\pgfpathlineto{\pgfqpoint{0.024056in}{-0.024056in}}%
\pgfpathlineto{\pgfqpoint{0.024056in}{0.024056in}}%
\pgfpathlineto{\pgfqpoint{-0.024056in}{0.024056in}}%
\pgfpathlineto{\pgfqpoint{-0.024056in}{-0.024056in}}%
\pgfpathclose%
\pgfusepath{stroke,fill}%
}%
\begin{pgfscope}%
\pgfsys@transformshift{2.671534in}{2.189570in}%
\pgfsys@useobject{currentmarker}{}%
\end{pgfscope}%
\begin{pgfscope}%
\pgfsys@transformshift{1.708223in}{1.679869in}%
\pgfsys@useobject{currentmarker}{}%
\end{pgfscope}%
\begin{pgfscope}%
\pgfsys@transformshift{2.234944in}{1.741278in}%
\pgfsys@useobject{currentmarker}{}%
\end{pgfscope}%
\begin{pgfscope}%
\pgfsys@transformshift{1.546903in}{1.686308in}%
\pgfsys@useobject{currentmarker}{}%
\end{pgfscope}%
\end{pgfscope}%
\begin{pgfscope}%
\pgfpathrectangle{\pgfqpoint{0.453704in}{0.415123in}}{\pgfqpoint{2.578386in}{1.858944in}}%
\pgfusepath{clip}%
\pgfsetbuttcap%
\pgfsetroundjoin%
\definecolor{currentfill}{rgb}{0.172549,0.627451,0.172549}%
\pgfsetfillcolor{currentfill}%
\pgfsetlinewidth{1.003750pt}%
\definecolor{currentstroke}{rgb}{0.172549,0.627451,0.172549}%
\pgfsetstrokecolor{currentstroke}%
\pgfsetdash{}{0pt}%
\pgfsys@defobject{currentmarker}{\pgfqpoint{-0.034021in}{-0.034021in}}{\pgfqpoint{0.034021in}{0.034021in}}{%
\pgfpathmoveto{\pgfqpoint{-0.000000in}{-0.034021in}}%
\pgfpathlineto{\pgfqpoint{0.034021in}{0.000000in}}%
\pgfpathlineto{\pgfqpoint{0.000000in}{0.034021in}}%
\pgfpathlineto{\pgfqpoint{-0.034021in}{0.000000in}}%
\pgfpathlineto{\pgfqpoint{-0.000000in}{-0.034021in}}%
\pgfpathclose%
\pgfusepath{stroke,fill}%
}%
\begin{pgfscope}%
\pgfsys@transformshift{1.212580in}{0.619300in}%
\pgfsys@useobject{currentmarker}{}%
\end{pgfscope}%
\begin{pgfscope}%
\pgfsys@transformshift{2.914891in}{0.854640in}%
\pgfsys@useobject{currentmarker}{}%
\end{pgfscope}%
\begin{pgfscope}%
\pgfsys@transformshift{1.472704in}{0.499621in}%
\pgfsys@useobject{currentmarker}{}%
\end{pgfscope}%
\begin{pgfscope}%
\pgfsys@transformshift{1.468977in}{0.844961in}%
\pgfsys@useobject{currentmarker}{}%
\end{pgfscope}%
\begin{pgfscope}%
\pgfsys@transformshift{2.589669in}{0.820349in}%
\pgfsys@useobject{currentmarker}{}%
\end{pgfscope}%
\begin{pgfscope}%
\pgfsys@transformshift{2.111425in}{0.802912in}%
\pgfsys@useobject{currentmarker}{}%
\end{pgfscope}%
\end{pgfscope}%
\begin{pgfscope}%
\pgfpathrectangle{\pgfqpoint{0.453704in}{0.415123in}}{\pgfqpoint{2.578386in}{1.858944in}}%
\pgfusepath{clip}%
\pgfsetbuttcap%
\pgfsetroundjoin%
\definecolor{currentfill}{rgb}{0.839216,0.152941,0.156863}%
\pgfsetfillcolor{currentfill}%
\pgfsetlinewidth{1.003750pt}%
\definecolor{currentstroke}{rgb}{0.839216,0.152941,0.156863}%
\pgfsetstrokecolor{currentstroke}%
\pgfsetdash{}{0pt}%
\pgfsys@defobject{currentmarker}{\pgfqpoint{-0.024056in}{-0.024056in}}{\pgfqpoint{0.024056in}{0.024056in}}{%
\pgfpathmoveto{\pgfqpoint{0.000000in}{0.024056in}}%
\pgfpathlineto{\pgfqpoint{-0.024056in}{-0.024056in}}%
\pgfpathlineto{\pgfqpoint{0.024056in}{-0.024056in}}%
\pgfpathlineto{\pgfqpoint{0.000000in}{0.024056in}}%
\pgfpathclose%
\pgfusepath{stroke,fill}%
}%
\begin{pgfscope}%
\pgfsys@transformshift{1.469990in}{1.582758in}%
\pgfsys@useobject{currentmarker}{}%
\end{pgfscope}%
\begin{pgfscope}%
\pgfsys@transformshift{2.112351in}{1.718199in}%
\pgfsys@useobject{currentmarker}{}%
\end{pgfscope}%
\end{pgfscope}%
\begin{pgfscope}%
\pgfsetbuttcap%
\pgfsetroundjoin%
\definecolor{currentfill}{rgb}{0.000000,0.000000,0.000000}%
\pgfsetfillcolor{currentfill}%
\pgfsetlinewidth{0.803000pt}%
\definecolor{currentstroke}{rgb}{0.000000,0.000000,0.000000}%
\pgfsetstrokecolor{currentstroke}%
\pgfsetdash{}{0pt}%
\pgfsys@defobject{currentmarker}{\pgfqpoint{0.000000in}{-0.048611in}}{\pgfqpoint{0.000000in}{0.000000in}}{%
\pgfpathmoveto{\pgfqpoint{0.000000in}{0.000000in}}%
\pgfpathlineto{\pgfqpoint{0.000000in}{-0.048611in}}%
\pgfusepath{stroke,fill}%
}%
\begin{pgfscope}%
\pgfsys@transformshift{1.334913in}{0.415123in}%
\pgfsys@useobject{currentmarker}{}%
\end{pgfscope}%
\end{pgfscope}%
\begin{pgfscope}%
\definecolor{textcolor}{rgb}{0.000000,0.000000,0.000000}%
\pgfsetstrokecolor{textcolor}%
\pgfsetfillcolor{textcolor}%
\pgftext[x=1.334913in,y=0.317901in,,top]{\color{textcolor}\rmfamily\fontsize{10.000000}{12.000000}\selectfont \(\displaystyle {10^{8}}\)}%
\end{pgfscope}%
\begin{pgfscope}%
\pgfsetbuttcap%
\pgfsetroundjoin%
\definecolor{currentfill}{rgb}{0.000000,0.000000,0.000000}%
\pgfsetfillcolor{currentfill}%
\pgfsetlinewidth{0.803000pt}%
\definecolor{currentstroke}{rgb}{0.000000,0.000000,0.000000}%
\pgfsetstrokecolor{currentstroke}%
\pgfsetdash{}{0pt}%
\pgfsys@defobject{currentmarker}{\pgfqpoint{0.000000in}{-0.048611in}}{\pgfqpoint{0.000000in}{0.000000in}}{%
\pgfpathmoveto{\pgfqpoint{0.000000in}{0.000000in}}%
\pgfpathlineto{\pgfqpoint{0.000000in}{-0.048611in}}%
\pgfusepath{stroke,fill}%
}%
\begin{pgfscope}%
\pgfsys@transformshift{2.746722in}{0.415123in}%
\pgfsys@useobject{currentmarker}{}%
\end{pgfscope}%
\end{pgfscope}%
\begin{pgfscope}%
\definecolor{textcolor}{rgb}{0.000000,0.000000,0.000000}%
\pgfsetstrokecolor{textcolor}%
\pgfsetfillcolor{textcolor}%
\pgftext[x=2.746722in,y=0.317901in,,top]{\color{textcolor}\rmfamily\fontsize{10.000000}{12.000000}\selectfont \(\displaystyle {10^{9}}\)}%
\end{pgfscope}%
\begin{pgfscope}%
\pgfsetbuttcap%
\pgfsetroundjoin%
\definecolor{currentfill}{rgb}{0.000000,0.000000,0.000000}%
\pgfsetfillcolor{currentfill}%
\pgfsetlinewidth{0.602250pt}%
\definecolor{currentstroke}{rgb}{0.000000,0.000000,0.000000}%
\pgfsetstrokecolor{currentstroke}%
\pgfsetdash{}{0pt}%
\pgfsys@defobject{currentmarker}{\pgfqpoint{0.000000in}{-0.027778in}}{\pgfqpoint{0.000000in}{0.000000in}}{%
\pgfpathmoveto{\pgfqpoint{0.000000in}{0.000000in}}%
\pgfpathlineto{\pgfqpoint{0.000000in}{-0.027778in}}%
\pgfusepath{stroke,fill}%
}%
\begin{pgfscope}%
\pgfsys@transformshift{0.596708in}{0.415123in}%
\pgfsys@useobject{currentmarker}{}%
\end{pgfscope}%
\end{pgfscope}%
\begin{pgfscope}%
\pgfsetbuttcap%
\pgfsetroundjoin%
\definecolor{currentfill}{rgb}{0.000000,0.000000,0.000000}%
\pgfsetfillcolor{currentfill}%
\pgfsetlinewidth{0.602250pt}%
\definecolor{currentstroke}{rgb}{0.000000,0.000000,0.000000}%
\pgfsetstrokecolor{currentstroke}%
\pgfsetdash{}{0pt}%
\pgfsys@defobject{currentmarker}{\pgfqpoint{0.000000in}{-0.027778in}}{\pgfqpoint{0.000000in}{0.000000in}}{%
\pgfpathmoveto{\pgfqpoint{0.000000in}{0.000000in}}%
\pgfpathlineto{\pgfqpoint{0.000000in}{-0.027778in}}%
\pgfusepath{stroke,fill}%
}%
\begin{pgfscope}%
\pgfsys@transformshift{0.773097in}{0.415123in}%
\pgfsys@useobject{currentmarker}{}%
\end{pgfscope}%
\end{pgfscope}%
\begin{pgfscope}%
\pgfsetbuttcap%
\pgfsetroundjoin%
\definecolor{currentfill}{rgb}{0.000000,0.000000,0.000000}%
\pgfsetfillcolor{currentfill}%
\pgfsetlinewidth{0.602250pt}%
\definecolor{currentstroke}{rgb}{0.000000,0.000000,0.000000}%
\pgfsetstrokecolor{currentstroke}%
\pgfsetdash{}{0pt}%
\pgfsys@defobject{currentmarker}{\pgfqpoint{0.000000in}{-0.027778in}}{\pgfqpoint{0.000000in}{0.000000in}}{%
\pgfpathmoveto{\pgfqpoint{0.000000in}{0.000000in}}%
\pgfpathlineto{\pgfqpoint{0.000000in}{-0.027778in}}%
\pgfusepath{stroke,fill}%
}%
\begin{pgfscope}%
\pgfsys@transformshift{0.909916in}{0.415123in}%
\pgfsys@useobject{currentmarker}{}%
\end{pgfscope}%
\end{pgfscope}%
\begin{pgfscope}%
\pgfsetbuttcap%
\pgfsetroundjoin%
\definecolor{currentfill}{rgb}{0.000000,0.000000,0.000000}%
\pgfsetfillcolor{currentfill}%
\pgfsetlinewidth{0.602250pt}%
\definecolor{currentstroke}{rgb}{0.000000,0.000000,0.000000}%
\pgfsetstrokecolor{currentstroke}%
\pgfsetdash{}{0pt}%
\pgfsys@defobject{currentmarker}{\pgfqpoint{0.000000in}{-0.027778in}}{\pgfqpoint{0.000000in}{0.000000in}}{%
\pgfpathmoveto{\pgfqpoint{0.000000in}{0.000000in}}%
\pgfpathlineto{\pgfqpoint{0.000000in}{-0.027778in}}%
\pgfusepath{stroke,fill}%
}%
\begin{pgfscope}%
\pgfsys@transformshift{1.021704in}{0.415123in}%
\pgfsys@useobject{currentmarker}{}%
\end{pgfscope}%
\end{pgfscope}%
\begin{pgfscope}%
\pgfsetbuttcap%
\pgfsetroundjoin%
\definecolor{currentfill}{rgb}{0.000000,0.000000,0.000000}%
\pgfsetfillcolor{currentfill}%
\pgfsetlinewidth{0.602250pt}%
\definecolor{currentstroke}{rgb}{0.000000,0.000000,0.000000}%
\pgfsetstrokecolor{currentstroke}%
\pgfsetdash{}{0pt}%
\pgfsys@defobject{currentmarker}{\pgfqpoint{0.000000in}{-0.027778in}}{\pgfqpoint{0.000000in}{0.000000in}}{%
\pgfpathmoveto{\pgfqpoint{0.000000in}{0.000000in}}%
\pgfpathlineto{\pgfqpoint{0.000000in}{-0.027778in}}%
\pgfusepath{stroke,fill}%
}%
\begin{pgfscope}%
\pgfsys@transformshift{1.116221in}{0.415123in}%
\pgfsys@useobject{currentmarker}{}%
\end{pgfscope}%
\end{pgfscope}%
\begin{pgfscope}%
\pgfsetbuttcap%
\pgfsetroundjoin%
\definecolor{currentfill}{rgb}{0.000000,0.000000,0.000000}%
\pgfsetfillcolor{currentfill}%
\pgfsetlinewidth{0.602250pt}%
\definecolor{currentstroke}{rgb}{0.000000,0.000000,0.000000}%
\pgfsetstrokecolor{currentstroke}%
\pgfsetdash{}{0pt}%
\pgfsys@defobject{currentmarker}{\pgfqpoint{0.000000in}{-0.027778in}}{\pgfqpoint{0.000000in}{0.000000in}}{%
\pgfpathmoveto{\pgfqpoint{0.000000in}{0.000000in}}%
\pgfpathlineto{\pgfqpoint{0.000000in}{-0.027778in}}%
\pgfusepath{stroke,fill}%
}%
\begin{pgfscope}%
\pgfsys@transformshift{1.198094in}{0.415123in}%
\pgfsys@useobject{currentmarker}{}%
\end{pgfscope}%
\end{pgfscope}%
\begin{pgfscope}%
\pgfsetbuttcap%
\pgfsetroundjoin%
\definecolor{currentfill}{rgb}{0.000000,0.000000,0.000000}%
\pgfsetfillcolor{currentfill}%
\pgfsetlinewidth{0.602250pt}%
\definecolor{currentstroke}{rgb}{0.000000,0.000000,0.000000}%
\pgfsetstrokecolor{currentstroke}%
\pgfsetdash{}{0pt}%
\pgfsys@defobject{currentmarker}{\pgfqpoint{0.000000in}{-0.027778in}}{\pgfqpoint{0.000000in}{0.000000in}}{%
\pgfpathmoveto{\pgfqpoint{0.000000in}{0.000000in}}%
\pgfpathlineto{\pgfqpoint{0.000000in}{-0.027778in}}%
\pgfusepath{stroke,fill}%
}%
\begin{pgfscope}%
\pgfsys@transformshift{1.270312in}{0.415123in}%
\pgfsys@useobject{currentmarker}{}%
\end{pgfscope}%
\end{pgfscope}%
\begin{pgfscope}%
\pgfsetbuttcap%
\pgfsetroundjoin%
\definecolor{currentfill}{rgb}{0.000000,0.000000,0.000000}%
\pgfsetfillcolor{currentfill}%
\pgfsetlinewidth{0.602250pt}%
\definecolor{currentstroke}{rgb}{0.000000,0.000000,0.000000}%
\pgfsetstrokecolor{currentstroke}%
\pgfsetdash{}{0pt}%
\pgfsys@defobject{currentmarker}{\pgfqpoint{0.000000in}{-0.027778in}}{\pgfqpoint{0.000000in}{0.000000in}}{%
\pgfpathmoveto{\pgfqpoint{0.000000in}{0.000000in}}%
\pgfpathlineto{\pgfqpoint{0.000000in}{-0.027778in}}%
\pgfusepath{stroke,fill}%
}%
\begin{pgfscope}%
\pgfsys@transformshift{1.759909in}{0.415123in}%
\pgfsys@useobject{currentmarker}{}%
\end{pgfscope}%
\end{pgfscope}%
\begin{pgfscope}%
\pgfsetbuttcap%
\pgfsetroundjoin%
\definecolor{currentfill}{rgb}{0.000000,0.000000,0.000000}%
\pgfsetfillcolor{currentfill}%
\pgfsetlinewidth{0.602250pt}%
\definecolor{currentstroke}{rgb}{0.000000,0.000000,0.000000}%
\pgfsetstrokecolor{currentstroke}%
\pgfsetdash{}{0pt}%
\pgfsys@defobject{currentmarker}{\pgfqpoint{0.000000in}{-0.027778in}}{\pgfqpoint{0.000000in}{0.000000in}}{%
\pgfpathmoveto{\pgfqpoint{0.000000in}{0.000000in}}%
\pgfpathlineto{\pgfqpoint{0.000000in}{-0.027778in}}%
\pgfusepath{stroke,fill}%
}%
\begin{pgfscope}%
\pgfsys@transformshift{2.008517in}{0.415123in}%
\pgfsys@useobject{currentmarker}{}%
\end{pgfscope}%
\end{pgfscope}%
\begin{pgfscope}%
\pgfsetbuttcap%
\pgfsetroundjoin%
\definecolor{currentfill}{rgb}{0.000000,0.000000,0.000000}%
\pgfsetfillcolor{currentfill}%
\pgfsetlinewidth{0.602250pt}%
\definecolor{currentstroke}{rgb}{0.000000,0.000000,0.000000}%
\pgfsetstrokecolor{currentstroke}%
\pgfsetdash{}{0pt}%
\pgfsys@defobject{currentmarker}{\pgfqpoint{0.000000in}{-0.027778in}}{\pgfqpoint{0.000000in}{0.000000in}}{%
\pgfpathmoveto{\pgfqpoint{0.000000in}{0.000000in}}%
\pgfpathlineto{\pgfqpoint{0.000000in}{-0.027778in}}%
\pgfusepath{stroke,fill}%
}%
\begin{pgfscope}%
\pgfsys@transformshift{2.184906in}{0.415123in}%
\pgfsys@useobject{currentmarker}{}%
\end{pgfscope}%
\end{pgfscope}%
\begin{pgfscope}%
\pgfsetbuttcap%
\pgfsetroundjoin%
\definecolor{currentfill}{rgb}{0.000000,0.000000,0.000000}%
\pgfsetfillcolor{currentfill}%
\pgfsetlinewidth{0.602250pt}%
\definecolor{currentstroke}{rgb}{0.000000,0.000000,0.000000}%
\pgfsetstrokecolor{currentstroke}%
\pgfsetdash{}{0pt}%
\pgfsys@defobject{currentmarker}{\pgfqpoint{0.000000in}{-0.027778in}}{\pgfqpoint{0.000000in}{0.000000in}}{%
\pgfpathmoveto{\pgfqpoint{0.000000in}{0.000000in}}%
\pgfpathlineto{\pgfqpoint{0.000000in}{-0.027778in}}%
\pgfusepath{stroke,fill}%
}%
\begin{pgfscope}%
\pgfsys@transformshift{2.321725in}{0.415123in}%
\pgfsys@useobject{currentmarker}{}%
\end{pgfscope}%
\end{pgfscope}%
\begin{pgfscope}%
\pgfsetbuttcap%
\pgfsetroundjoin%
\definecolor{currentfill}{rgb}{0.000000,0.000000,0.000000}%
\pgfsetfillcolor{currentfill}%
\pgfsetlinewidth{0.602250pt}%
\definecolor{currentstroke}{rgb}{0.000000,0.000000,0.000000}%
\pgfsetstrokecolor{currentstroke}%
\pgfsetdash{}{0pt}%
\pgfsys@defobject{currentmarker}{\pgfqpoint{0.000000in}{-0.027778in}}{\pgfqpoint{0.000000in}{0.000000in}}{%
\pgfpathmoveto{\pgfqpoint{0.000000in}{0.000000in}}%
\pgfpathlineto{\pgfqpoint{0.000000in}{-0.027778in}}%
\pgfusepath{stroke,fill}%
}%
\begin{pgfscope}%
\pgfsys@transformshift{2.433514in}{0.415123in}%
\pgfsys@useobject{currentmarker}{}%
\end{pgfscope}%
\end{pgfscope}%
\begin{pgfscope}%
\pgfsetbuttcap%
\pgfsetroundjoin%
\definecolor{currentfill}{rgb}{0.000000,0.000000,0.000000}%
\pgfsetfillcolor{currentfill}%
\pgfsetlinewidth{0.602250pt}%
\definecolor{currentstroke}{rgb}{0.000000,0.000000,0.000000}%
\pgfsetstrokecolor{currentstroke}%
\pgfsetdash{}{0pt}%
\pgfsys@defobject{currentmarker}{\pgfqpoint{0.000000in}{-0.027778in}}{\pgfqpoint{0.000000in}{0.000000in}}{%
\pgfpathmoveto{\pgfqpoint{0.000000in}{0.000000in}}%
\pgfpathlineto{\pgfqpoint{0.000000in}{-0.027778in}}%
\pgfusepath{stroke,fill}%
}%
\begin{pgfscope}%
\pgfsys@transformshift{2.528030in}{0.415123in}%
\pgfsys@useobject{currentmarker}{}%
\end{pgfscope}%
\end{pgfscope}%
\begin{pgfscope}%
\pgfsetbuttcap%
\pgfsetroundjoin%
\definecolor{currentfill}{rgb}{0.000000,0.000000,0.000000}%
\pgfsetfillcolor{currentfill}%
\pgfsetlinewidth{0.602250pt}%
\definecolor{currentstroke}{rgb}{0.000000,0.000000,0.000000}%
\pgfsetstrokecolor{currentstroke}%
\pgfsetdash{}{0pt}%
\pgfsys@defobject{currentmarker}{\pgfqpoint{0.000000in}{-0.027778in}}{\pgfqpoint{0.000000in}{0.000000in}}{%
\pgfpathmoveto{\pgfqpoint{0.000000in}{0.000000in}}%
\pgfpathlineto{\pgfqpoint{0.000000in}{-0.027778in}}%
\pgfusepath{stroke,fill}%
}%
\begin{pgfscope}%
\pgfsys@transformshift{2.609903in}{0.415123in}%
\pgfsys@useobject{currentmarker}{}%
\end{pgfscope}%
\end{pgfscope}%
\begin{pgfscope}%
\pgfsetbuttcap%
\pgfsetroundjoin%
\definecolor{currentfill}{rgb}{0.000000,0.000000,0.000000}%
\pgfsetfillcolor{currentfill}%
\pgfsetlinewidth{0.602250pt}%
\definecolor{currentstroke}{rgb}{0.000000,0.000000,0.000000}%
\pgfsetstrokecolor{currentstroke}%
\pgfsetdash{}{0pt}%
\pgfsys@defobject{currentmarker}{\pgfqpoint{0.000000in}{-0.027778in}}{\pgfqpoint{0.000000in}{0.000000in}}{%
\pgfpathmoveto{\pgfqpoint{0.000000in}{0.000000in}}%
\pgfpathlineto{\pgfqpoint{0.000000in}{-0.027778in}}%
\pgfusepath{stroke,fill}%
}%
\begin{pgfscope}%
\pgfsys@transformshift{2.682121in}{0.415123in}%
\pgfsys@useobject{currentmarker}{}%
\end{pgfscope}%
\end{pgfscope}%
\begin{pgfscope}%
\definecolor{textcolor}{rgb}{0.000000,0.000000,0.000000}%
\pgfsetstrokecolor{textcolor}%
\pgfsetfillcolor{textcolor}%
\pgftext[x=1.742897in,y=0.138889in,,top]{\color{textcolor}\rmfamily\fontsize{10.000000}{12.000000}\selectfont Model Size (\# Parameters)}%
\end{pgfscope}%
\begin{pgfscope}%
\pgfsetbuttcap%
\pgfsetroundjoin%
\definecolor{currentfill}{rgb}{0.000000,0.000000,0.000000}%
\pgfsetfillcolor{currentfill}%
\pgfsetlinewidth{0.803000pt}%
\definecolor{currentstroke}{rgb}{0.000000,0.000000,0.000000}%
\pgfsetstrokecolor{currentstroke}%
\pgfsetdash{}{0pt}%
\pgfsys@defobject{currentmarker}{\pgfqpoint{-0.048611in}{0.000000in}}{\pgfqpoint{-0.000000in}{0.000000in}}{%
\pgfpathmoveto{\pgfqpoint{-0.000000in}{0.000000in}}%
\pgfpathlineto{\pgfqpoint{-0.048611in}{0.000000in}}%
\pgfusepath{stroke,fill}%
}%
\begin{pgfscope}%
\pgfsys@transformshift{0.453704in}{0.802912in}%
\pgfsys@useobject{currentmarker}{}%
\end{pgfscope}%
\end{pgfscope}%
\begin{pgfscope}%
\definecolor{textcolor}{rgb}{0.000000,0.000000,0.000000}%
\pgfsetstrokecolor{textcolor}%
\pgfsetfillcolor{textcolor}%
\pgftext[x=0.179012in, y=0.754687in, left, base]{\color{textcolor}\rmfamily\fontsize{10.000000}{12.000000}\selectfont \(\displaystyle {0.5}\)}%
\end{pgfscope}%
\begin{pgfscope}%
\pgfsetbuttcap%
\pgfsetroundjoin%
\definecolor{currentfill}{rgb}{0.000000,0.000000,0.000000}%
\pgfsetfillcolor{currentfill}%
\pgfsetlinewidth{0.803000pt}%
\definecolor{currentstroke}{rgb}{0.000000,0.000000,0.000000}%
\pgfsetstrokecolor{currentstroke}%
\pgfsetdash{}{0pt}%
\pgfsys@defobject{currentmarker}{\pgfqpoint{-0.048611in}{0.000000in}}{\pgfqpoint{-0.000000in}{0.000000in}}{%
\pgfpathmoveto{\pgfqpoint{-0.000000in}{0.000000in}}%
\pgfpathlineto{\pgfqpoint{-0.048611in}{0.000000in}}%
\pgfusepath{stroke,fill}%
}%
\begin{pgfscope}%
\pgfsys@transformshift{0.453704in}{1.420198in}%
\pgfsys@useobject{currentmarker}{}%
\end{pgfscope}%
\end{pgfscope}%
\begin{pgfscope}%
\definecolor{textcolor}{rgb}{0.000000,0.000000,0.000000}%
\pgfsetstrokecolor{textcolor}%
\pgfsetfillcolor{textcolor}%
\pgftext[x=0.179012in, y=1.371973in, left, base]{\color{textcolor}\rmfamily\fontsize{10.000000}{12.000000}\selectfont \(\displaystyle {0.6}\)}%
\end{pgfscope}%
\begin{pgfscope}%
\pgfsetbuttcap%
\pgfsetroundjoin%
\definecolor{currentfill}{rgb}{0.000000,0.000000,0.000000}%
\pgfsetfillcolor{currentfill}%
\pgfsetlinewidth{0.803000pt}%
\definecolor{currentstroke}{rgb}{0.000000,0.000000,0.000000}%
\pgfsetstrokecolor{currentstroke}%
\pgfsetdash{}{0pt}%
\pgfsys@defobject{currentmarker}{\pgfqpoint{-0.048611in}{0.000000in}}{\pgfqpoint{-0.000000in}{0.000000in}}{%
\pgfpathmoveto{\pgfqpoint{-0.000000in}{0.000000in}}%
\pgfpathlineto{\pgfqpoint{-0.048611in}{0.000000in}}%
\pgfusepath{stroke,fill}%
}%
\begin{pgfscope}%
\pgfsys@transformshift{0.453704in}{2.037485in}%
\pgfsys@useobject{currentmarker}{}%
\end{pgfscope}%
\end{pgfscope}%
\begin{pgfscope}%
\definecolor{textcolor}{rgb}{0.000000,0.000000,0.000000}%
\pgfsetstrokecolor{textcolor}%
\pgfsetfillcolor{textcolor}%
\pgftext[x=0.179012in, y=1.989260in, left, base]{\color{textcolor}\rmfamily\fontsize{10.000000}{12.000000}\selectfont \(\displaystyle {0.7}\)}%
\end{pgfscope}%
\begin{pgfscope}%
\pgfsetbuttcap%
\pgfsetroundjoin%
\definecolor{currentfill}{rgb}{0.000000,0.000000,0.000000}%
\pgfsetfillcolor{currentfill}%
\pgfsetlinewidth{0.602250pt}%
\definecolor{currentstroke}{rgb}{0.000000,0.000000,0.000000}%
\pgfsetstrokecolor{currentstroke}%
\pgfsetdash{}{0pt}%
\pgfsys@defobject{currentmarker}{\pgfqpoint{-0.027778in}{0.000000in}}{\pgfqpoint{-0.000000in}{0.000000in}}{%
\pgfpathmoveto{\pgfqpoint{-0.000000in}{0.000000in}}%
\pgfpathlineto{\pgfqpoint{-0.027778in}{0.000000in}}%
\pgfusepath{stroke,fill}%
}%
\begin{pgfscope}%
\pgfsys@transformshift{0.453704in}{0.494269in}%
\pgfsys@useobject{currentmarker}{}%
\end{pgfscope}%
\end{pgfscope}%
\begin{pgfscope}%
\pgfsetbuttcap%
\pgfsetroundjoin%
\definecolor{currentfill}{rgb}{0.000000,0.000000,0.000000}%
\pgfsetfillcolor{currentfill}%
\pgfsetlinewidth{0.602250pt}%
\definecolor{currentstroke}{rgb}{0.000000,0.000000,0.000000}%
\pgfsetstrokecolor{currentstroke}%
\pgfsetdash{}{0pt}%
\pgfsys@defobject{currentmarker}{\pgfqpoint{-0.027778in}{0.000000in}}{\pgfqpoint{-0.000000in}{0.000000in}}{%
\pgfpathmoveto{\pgfqpoint{-0.000000in}{0.000000in}}%
\pgfpathlineto{\pgfqpoint{-0.027778in}{0.000000in}}%
\pgfusepath{stroke,fill}%
}%
\begin{pgfscope}%
\pgfsys@transformshift{0.453704in}{1.111555in}%
\pgfsys@useobject{currentmarker}{}%
\end{pgfscope}%
\end{pgfscope}%
\begin{pgfscope}%
\pgfsetbuttcap%
\pgfsetroundjoin%
\definecolor{currentfill}{rgb}{0.000000,0.000000,0.000000}%
\pgfsetfillcolor{currentfill}%
\pgfsetlinewidth{0.602250pt}%
\definecolor{currentstroke}{rgb}{0.000000,0.000000,0.000000}%
\pgfsetstrokecolor{currentstroke}%
\pgfsetdash{}{0pt}%
\pgfsys@defobject{currentmarker}{\pgfqpoint{-0.027778in}{0.000000in}}{\pgfqpoint{-0.000000in}{0.000000in}}{%
\pgfpathmoveto{\pgfqpoint{-0.000000in}{0.000000in}}%
\pgfpathlineto{\pgfqpoint{-0.027778in}{0.000000in}}%
\pgfusepath{stroke,fill}%
}%
\begin{pgfscope}%
\pgfsys@transformshift{0.453704in}{1.728842in}%
\pgfsys@useobject{currentmarker}{}%
\end{pgfscope}%
\end{pgfscope}%
\begin{pgfscope}%
\definecolor{textcolor}{rgb}{0.000000,0.000000,0.000000}%
\pgfsetstrokecolor{textcolor}%
\pgfsetfillcolor{textcolor}%
\pgftext[x=0.123457in,y=1.344595in,,bottom,rotate=90.000000]{\color{textcolor}\rmfamily\fontsize{10.000000}{12.000000}\selectfont Error Detection F1}%
\end{pgfscope}%
\begin{pgfscope}%
\pgfpathrectangle{\pgfqpoint{0.453704in}{0.415123in}}{\pgfqpoint{2.578386in}{1.858944in}}%
\pgfusepath{clip}%
\pgfsetbuttcap%
\pgfsetroundjoin%
\pgfsetlinewidth{1.505625pt}%
\definecolor{currentstroke}{rgb}{0.121569,0.466667,0.705882}%
\pgfsetstrokecolor{currentstroke}%
\pgfsetdash{{5.550000pt}{2.400000pt}}{0.000000pt}%
\pgfpathmoveto{\pgfqpoint{1.390458in}{1.532341in}}%
\pgfpathlineto{\pgfqpoint{2.073570in}{1.603774in}}%
\pgfpathlineto{\pgfqpoint{0.570904in}{1.446639in}}%
\pgfpathlineto{\pgfqpoint{1.083505in}{1.500242in}}%
\pgfusepath{stroke}%
\end{pgfscope}%
\begin{pgfscope}%
\pgfpathrectangle{\pgfqpoint{0.453704in}{0.415123in}}{\pgfqpoint{2.578386in}{1.858944in}}%
\pgfusepath{clip}%
\pgfsetbuttcap%
\pgfsetroundjoin%
\pgfsetlinewidth{1.505625pt}%
\definecolor{currentstroke}{rgb}{1.000000,0.498039,0.054902}%
\pgfsetstrokecolor{currentstroke}%
\pgfsetdash{{5.550000pt}{2.400000pt}}{0.000000pt}%
\pgfpathmoveto{\pgfqpoint{2.671534in}{2.088241in}}%
\pgfpathlineto{\pgfqpoint{1.708223in}{1.685316in}}%
\pgfpathlineto{\pgfqpoint{2.234944in}{1.905628in}}%
\pgfpathlineto{\pgfqpoint{1.546903in}{1.617840in}}%
\pgfusepath{stroke}%
\end{pgfscope}%
\begin{pgfscope}%
\pgfpathrectangle{\pgfqpoint{0.453704in}{0.415123in}}{\pgfqpoint{2.578386in}{1.858944in}}%
\pgfusepath{clip}%
\pgfsetbuttcap%
\pgfsetroundjoin%
\pgfsetlinewidth{1.505625pt}%
\definecolor{currentstroke}{rgb}{0.172549,0.627451,0.172549}%
\pgfsetstrokecolor{currentstroke}%
\pgfsetdash{{5.550000pt}{2.400000pt}}{0.000000pt}%
\pgfpathmoveto{\pgfqpoint{1.212580in}{0.637374in}}%
\pgfpathlineto{\pgfqpoint{2.914891in}{0.871255in}}%
\pgfpathlineto{\pgfqpoint{1.472704in}{0.673113in}}%
\pgfpathlineto{\pgfqpoint{1.468977in}{0.672601in}}%
\pgfpathlineto{\pgfqpoint{2.589669in}{0.826573in}}%
\pgfpathlineto{\pgfqpoint{2.111425in}{0.760867in}}%
\pgfusepath{stroke}%
\end{pgfscope}%
\begin{pgfscope}%
\pgfpathrectangle{\pgfqpoint{0.453704in}{0.415123in}}{\pgfqpoint{2.578386in}{1.858944in}}%
\pgfusepath{clip}%
\pgfsetbuttcap%
\pgfsetroundjoin%
\pgfsetlinewidth{1.505625pt}%
\definecolor{currentstroke}{rgb}{0.839216,0.152941,0.156863}%
\pgfsetstrokecolor{currentstroke}%
\pgfsetdash{{5.550000pt}{2.400000pt}}{0.000000pt}%
\pgfpathmoveto{\pgfqpoint{1.469990in}{1.582758in}}%
\pgfpathlineto{\pgfqpoint{2.112351in}{1.718199in}}%
\pgfusepath{stroke}%
\end{pgfscope}%
\begin{pgfscope}%
\pgfsetrectcap%
\pgfsetmiterjoin%
\pgfsetlinewidth{0.803000pt}%
\definecolor{currentstroke}{rgb}{0.000000,0.000000,0.000000}%
\pgfsetstrokecolor{currentstroke}%
\pgfsetdash{}{0pt}%
\pgfpathmoveto{\pgfqpoint{0.453704in}{0.415123in}}%
\pgfpathlineto{\pgfqpoint{0.453704in}{2.274068in}}%
\pgfusepath{stroke}%
\end{pgfscope}%
\begin{pgfscope}%
\pgfsetrectcap%
\pgfsetmiterjoin%
\pgfsetlinewidth{0.803000pt}%
\definecolor{currentstroke}{rgb}{0.000000,0.000000,0.000000}%
\pgfsetstrokecolor{currentstroke}%
\pgfsetdash{}{0pt}%
\pgfpathmoveto{\pgfqpoint{3.032090in}{0.415123in}}%
\pgfpathlineto{\pgfqpoint{3.032090in}{2.274068in}}%
\pgfusepath{stroke}%
\end{pgfscope}%
\begin{pgfscope}%
\pgfsetrectcap%
\pgfsetmiterjoin%
\pgfsetlinewidth{0.803000pt}%
\definecolor{currentstroke}{rgb}{0.000000,0.000000,0.000000}%
\pgfsetstrokecolor{currentstroke}%
\pgfsetdash{}{0pt}%
\pgfpathmoveto{\pgfqpoint{0.453704in}{0.415123in}}%
\pgfpathlineto{\pgfqpoint{3.032090in}{0.415123in}}%
\pgfusepath{stroke}%
\end{pgfscope}%
\begin{pgfscope}%
\pgfsetrectcap%
\pgfsetmiterjoin%
\pgfsetlinewidth{0.803000pt}%
\definecolor{currentstroke}{rgb}{0.000000,0.000000,0.000000}%
\pgfsetstrokecolor{currentstroke}%
\pgfsetdash{}{0pt}%
\pgfpathmoveto{\pgfqpoint{0.453704in}{2.274068in}}%
\pgfpathlineto{\pgfqpoint{3.032090in}{2.274068in}}%
\pgfusepath{stroke}%
\end{pgfscope}%
\begin{pgfscope}%
\pgfsetbuttcap%
\pgfsetmiterjoin%
\definecolor{currentfill}{rgb}{1.000000,1.000000,1.000000}%
\pgfsetfillcolor{currentfill}%
\pgfsetfillopacity{0.800000}%
\pgfsetlinewidth{1.003750pt}%
\definecolor{currentstroke}{rgb}{0.800000,0.800000,0.800000}%
\pgfsetstrokecolor{currentstroke}%
\pgfsetstrokeopacity{0.800000}%
\pgfsetdash{}{0pt}%
\pgfpathmoveto{\pgfqpoint{0.516899in}{1.952771in}}%
\pgfpathlineto{\pgfqpoint{2.125452in}{1.952771in}}%
\pgfpathquadraticcurveto{\pgfqpoint{2.143508in}{1.952771in}}{\pgfqpoint{2.143508in}{1.970827in}}%
\pgfpathlineto{\pgfqpoint{2.143508in}{2.210873in}}%
\pgfpathquadraticcurveto{\pgfqpoint{2.143508in}{2.228929in}}{\pgfqpoint{2.125452in}{2.228929in}}%
\pgfpathlineto{\pgfqpoint{0.516899in}{2.228929in}}%
\pgfpathquadraticcurveto{\pgfqpoint{0.498843in}{2.228929in}}{\pgfqpoint{0.498843in}{2.210873in}}%
\pgfpathlineto{\pgfqpoint{0.498843in}{1.970827in}}%
\pgfpathquadraticcurveto{\pgfqpoint{0.498843in}{1.952771in}}{\pgfqpoint{0.516899in}{1.952771in}}%
\pgfpathlineto{\pgfqpoint{0.516899in}{1.952771in}}%
\pgfpathclose%
\pgfusepath{stroke,fill}%
\end{pgfscope}%
\begin{pgfscope}%
\pgfsetbuttcap%
\pgfsetroundjoin%
\pgfsetlinewidth{1.505625pt}%
\definecolor{currentstroke}{rgb}{0.121569,0.466667,0.705882}%
\pgfsetstrokecolor{currentstroke}%
\pgfsetdash{{5.550000pt}{2.400000pt}}{0.000000pt}%
\pgfpathmoveto{\pgfqpoint{0.534954in}{2.161220in}}%
\pgfpathlineto{\pgfqpoint{0.625232in}{2.161220in}}%
\pgfpathlineto{\pgfqpoint{0.715510in}{2.161220in}}%
\pgfusepath{stroke}%
\end{pgfscope}%
\begin{pgfscope}%
\definecolor{textcolor}{rgb}{0.000000,0.000000,0.000000}%
\pgfsetstrokecolor{textcolor}%
\pgfsetfillcolor{textcolor}%
\pgftext[x=0.787732in,y=2.129623in,left,base]{\color{textcolor}\rmfamily\fontsize{6.500000}{7.800000}\selectfont BERT}%
\end{pgfscope}%
\begin{pgfscope}%
\pgfsetbuttcap%
\pgfsetroundjoin%
\pgfsetlinewidth{1.505625pt}%
\definecolor{currentstroke}{rgb}{1.000000,0.498039,0.054902}%
\pgfsetstrokecolor{currentstroke}%
\pgfsetdash{{5.550000pt}{2.400000pt}}{0.000000pt}%
\pgfpathmoveto{\pgfqpoint{0.534954in}{2.036683in}}%
\pgfpathlineto{\pgfqpoint{0.625232in}{2.036683in}}%
\pgfpathlineto{\pgfqpoint{0.715510in}{2.036683in}}%
\pgfusepath{stroke}%
\end{pgfscope}%
\begin{pgfscope}%
\definecolor{textcolor}{rgb}{0.000000,0.000000,0.000000}%
\pgfsetstrokecolor{textcolor}%
\pgfsetfillcolor{textcolor}%
\pgftext[x=0.787732in,y=2.005086in,left,base]{\color{textcolor}\rmfamily\fontsize{6.500000}{7.800000}\selectfont DeBERTa}%
\end{pgfscope}%
\begin{pgfscope}%
\pgfsetbuttcap%
\pgfsetroundjoin%
\pgfsetlinewidth{1.505625pt}%
\definecolor{currentstroke}{rgb}{0.172549,0.627451,0.172549}%
\pgfsetstrokecolor{currentstroke}%
\pgfsetdash{{5.550000pt}{2.400000pt}}{0.000000pt}%
\pgfpathmoveto{\pgfqpoint{1.410103in}{2.161220in}}%
\pgfpathlineto{\pgfqpoint{1.500381in}{2.161220in}}%
\pgfpathlineto{\pgfqpoint{1.590658in}{2.161220in}}%
\pgfusepath{stroke}%
\end{pgfscope}%
\begin{pgfscope}%
\definecolor{textcolor}{rgb}{0.000000,0.000000,0.000000}%
\pgfsetstrokecolor{textcolor}%
\pgfsetfillcolor{textcolor}%
\pgftext[x=1.662881in,y=2.129623in,left,base]{\color{textcolor}\rmfamily\fontsize{6.500000}{7.800000}\selectfont GPT}%
\end{pgfscope}%
\begin{pgfscope}%
\pgfsetbuttcap%
\pgfsetroundjoin%
\pgfsetlinewidth{1.505625pt}%
\definecolor{currentstroke}{rgb}{0.839216,0.152941,0.156863}%
\pgfsetstrokecolor{currentstroke}%
\pgfsetdash{{5.550000pt}{2.400000pt}}{0.000000pt}%
\pgfpathmoveto{\pgfqpoint{1.410103in}{2.036683in}}%
\pgfpathlineto{\pgfqpoint{1.500381in}{2.036683in}}%
\pgfpathlineto{\pgfqpoint{1.590658in}{2.036683in}}%
\pgfusepath{stroke}%
\end{pgfscope}%
\begin{pgfscope}%
\definecolor{textcolor}{rgb}{0.000000,0.000000,0.000000}%
\pgfsetstrokecolor{textcolor}%
\pgfsetfillcolor{textcolor}%
\pgftext[x=1.662881in,y=2.005086in,left,base]{\color{textcolor}\rmfamily\fontsize{6.500000}{7.800000}\selectfont RoBERTa}%
\end{pgfscope}%
\end{pgfpicture}%
\makeatother%
\endgroup%